\documentclass[a4paper,fleqn]{cas-dc}

\usepackage[authoryear,longnamesfirst]{natbib}
\usepackage{amssymb}
\usepackage{multirow}

\usepackage{amsmath}
\usepackage{algorithmic}
\usepackage{algorithm}
\usepackage{color}
\usepackage{makecell}
\usepackage{booktabs}
\usepackage[dvipsnames,svgnames]{xcolor}
\usepackage{tcolorbox}
\usepackage{hyperref}
\usepackage{rotating}

\newcommand{\blue}[1]{#1}

\def\tsc#1{\csdef{#1}{\textsc{\lowercase{#1}}\xspace}}
\tsc{WGM}
\tsc{QE}
\tsc{EP}
\tsc{PMS}
\tsc{BEC}
\tsc{DE}

\begin{document}
\let\WriteBookmarks\relax
\def\floatpagepagefraction{1}
\def\textpagefraction{.001}


\shortauthors{Gong et~al.}

\title [mode = title]{A Survey of Low-bit Large Language Models: Basics, Systems, and Algorithms}                      



%

\author[a]{Ruihao Gong}
\fnmark[1]

\author[a]{Yifu Ding}
\author[a]{Zining Wang}
\author[a]{Chengtao Lv}
\author[a]{Xingyu Zheng}
\author[a]{Jinyang Du}
\author[a]{Jinyang Guo}
\author[a]{Xianglong Liu}
\cormark[1]

\author[b]{Haotong Qin}
\author[b]{Michele Magno}

\author[c]{Yang Yong}
\author[c]{Shiqiao Gu}

\author[d,c]{Dahua Lin}

\affiliation[a]{organization={Beihang University},
    addressline={37 Xueyuan Road},
    city={Haidian District},
    postcode={100191},
    state={Beijing},
    country={China}}

\affiliation[b]{organization={ETH Zurich},
    addressline={Rämistrasse 101},
    city={Zurich},
    postcode={8092},
    state={Zurich},
    country={Switzerland}}

\affiliation[c]{organization={SenseTime},
    addressline={No.1900 Hongmei Road},
    city={Xuhui District},
    postcode={200233},
    state={Shanghai},
    country={China}}

\affiliation[d]{organization={The Chinese University of Hong Kong},
    addressline={Shatin},
    city={NT},
    postcode={999077},
    state={Hong Kong SAR},
    country={China}}

\cortext[cor1]{Corresponding author: Xianglong Liu (xlliu@buaa.edu.cn)}

\fntext[fn1]{Ruihao Gong leads the overall organization of the survey, with Yifu Ding and Jinyang Du contributing to Sections 2 and 3. 
Xingyu Zheng is responsible for authoring Section 4, while Chengtao Lv and Zining Wang collaborate on Section 5. 
Yang Yong and Shiqiao Gu contribute to the collection of toolkits and systems. 
Haotong Qin, Jinyang Guo, Dahua Lin, Michele Magno, and Xianglong Liu provide guidance throughout and assist in refining the final manuscript.}



\begin{abstract}
Large language models (LLMs) have achieved remarkable advancements in natural language processing, showcasing exceptional performance across various tasks. However, the expensive memory and computational requirements present significant challenges for their practical deployment. Low-bit quantization has emerged as a critical approach to mitigate these challenges by reducing the bit-width of model parameters, activations, and gradients, thus decreasing memory usage and computational demands. This paper presents a comprehensive survey of low-bit quantization methods tailored for LLMs, covering the fundamental principles, system implementations, and algorithmic strategies. An overview of basic concepts and new data formats specific to low-bit LLMs is first introduced, followed by a review of frameworks and systems that facilitate low-bit LLMs across various hardware platforms. Then, we categorize and analyze techniques and toolkits for efficient low-bit training and inference of LLMs. Finally, we conclude with a discussion of future trends and potential advancements of low-bit LLMs. 
Our systematic overview from basic, system, and algorithm perspectives can offer valuable insights and guidelines for future works to enhance the efficiency and applicability of LLMs through low-bit quantization.

\end{abstract}



\maketitle

\section{Introduction}
Large language models (LLMs)~\citep{openai2024gpt4, touvron2023llama, touvron2023llama2, dubey2024llama, lozhkov2024starcoder, liu2024deepseek} have revolutionized natural language processing by delivering unprecedented performance across a range of tasks, from text generation to language understanding. However, their remarkable capabilities come with significant computational and memory demands. This has raised considerable challenges when deploying these models in scenarios with limited resources or high concurrency. To address these challenges, low-bit quantization has emerged as a pivotal approach for enhancing the efficiency and deployability of LLMs. 

Low-bit quantization involves the process of reducing the bit-width of tensors, which effectively decreases the memory footprint and computational requirements of LLMs. 
By compressing weights, activations, and gradients of LLMs with low-bit integer/binary representation, quantization can significantly accelerate inference and training and reduce storage requirements with acceptable accuracy. 
This efficiency is crucial for enabling advanced LLMs to be accessible on devices with constrained resources, thereby broadening their applicability.

\begin{figure*}[h]
    \centering
    \includegraphics[width=\linewidth]{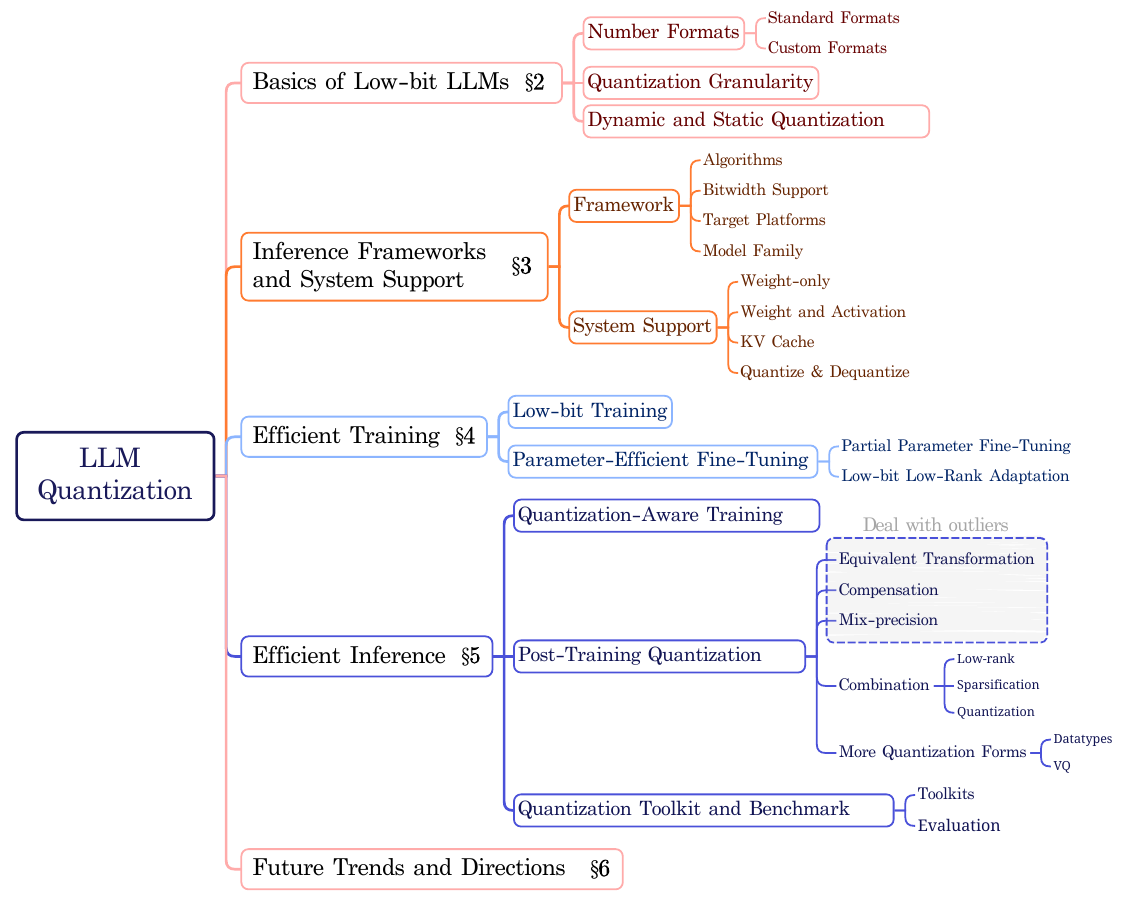}
    \caption{The skeleton of the LLM Quantization methods. The diagram illustrates the main areas in the survey. }
    \label{fig:overview}
\end{figure*}

In this paper, we aim to provide a survey with a comprehensive overview of low-bit quantization for large language models (LLMs), encompassing the fundamental concepts, system implementations, and algorithmic approaches related to low-bit LLMs. 
Compared with the traditional models, LLMs, as the representative paradigm of the foundation model, 
always feature a vast number of parameters, which presents unique challenges for effective quantization. As depicted in Figure~\ref{fig:overview}, Section~\ref{sec:basics} introduces the fundamentals of low-bit quantization of LLMs, including new low-bit data formats and quantization granularities specific to LLMs. Section~\ref{sec:system} reviews the systems and frameworks supporting low-bit LLMs across various hardware platforms. We then categorize low-bit quantization techniques for efficient training and inference in Sections~\ref{sec:training} and~\ref{sec:inference}, respectively. For training, we discuss methods for low-bit training and fine-tuning of LLMs. For inference, we differentiate LLM quantization methods by quantization-aware training and post-training quantization. Quantization-aware training is often used for low-bit settings (such as binary quantization). Post-training quantization is more commonly applied in existing research since it is a resource-efficient pipeline. For a clear understanding, we first cover the widely used techniques of equivalent transformation for reducing outlier influence and weight compensation for mitigating quantization errors. Then the mixed precision, techniques that combine quantization with other compression methods, as well as methods for new quantization forms are discussed. Additionally, we summarize toolkits that integrate these algorithms to support the development of accurate low-bit LLMs. Finally, Section~\ref{sec:future} explores future trends and directions, discussing emerging research areas, potential breakthroughs, and the impact of new technologies on LLM quantization. 
Our survey provides a detailed description of the fundamentals of low-bit LLMs and gives a comprehensive view of the system implementations for accelerating training and inference through low-bit quantization and algorithms and strategies to maintain and enhance quantized accuracy. We believe this survey can provide valuable insights and advance the development of LLM quantization.

\section{Basics of Low-bit LLMs}
\label{sec:basics}
In this section, we introduce the basic fundaments of quantization and low-bit LLMs from three aspects: (1) Low-bit number formats. To deal with the outliers in LLMs, low-bit floating-points are first used in quantization. And lots of custom data formats are designed to tackle the outliers. However, integers are still the mainstream. (2) Quantization granularity. To improve the performance of quantized LLMs, finer-grained quantization retains more information and generates better results. But course-grained ones occupy less storage and are more efficient in inference. (3) Dynamic or static quantization. Dynamic quantization does not require calibration, as the quantization parameters are calculated on the fly, making the preparation of a quantized model simpler. In contrast, static quantization requires pre-calibration of quantization parameters, but it offers faster inference performance.

\subsection{Low-bit Number Formats}

We start with the low-bit number formats at the beginning of the introduction. First, we demonstrate the standard formats that are well-recognized, but focus on the differences in LLMs. Second, we introduce some typical custom formats that are designed for LLMs. 

\subsubsection{Standard Formats}

\textbf{Floating-point Numbers}. 
The floating-point data type is comprehensively defined in the IEEE 754~\citep{ieee754} standard, which is also the most prevailing number format in computer systems. Let us denote them as $\mathrm{FP}k$, where $k$ represents the number of bits that the value occupies in memory, usually $32, 16, 8$, etc. A floating-point number can be uniformly expressed as:

\begin{eqnarray}
\label{eq:sec1:fp}
X_{\mathrm{FP}k}&=&{(-1)}^s2^{p-bias}(1.\texttt{mantissa})={(-1)}^s2^{p-bias}\nonumber\\
&&\left(1+\frac{d_1}{2}+\frac{d_2}{2^2}+\ldots+\frac{d_m}{2^m}\right),
\end{eqnarray}
where $s$ is the sign bit, $p$ is the exponent integer, $bias$ is applied to the exponent, $m$ is the total number of mantissa bits in the significand, and $d_1, d_2, \ldots, d_m$ represent the digits of the mantissa part in the binary format. 
The bits of $s$, $p$ and $m$ should be accumulated to $k$ for an $\mathrm{FP}k$ value.

Since LLMs occupy more memory, lower-bit formats become popularly adopted in both training and inference. We omit the 32-bit number format here since the 16 and lower bitwidth has become the mainstream practice in application. 
We can further categorize each $\mathrm{FP}k$ according to its bit allocations for the exponent (E) and mantissa (M) parts.  
We use $\mathrm{E}e\mathrm{M}m$ to denote the subcategories.
As for $\mathrm{FP}16$, IEEE 754 defines float16 (also known as half-precision or FP16) and bfloat16 (brain floating point or BF16), which can be represented as E5M10 and E8M7, respectively. Therefore, bfloat16 can represent larger magnitudes with more exponent bits (identical to that of FP32) while more sparse than float16 with less mantissa in the significand, which may exert unprecedented potential in LLMs~\citep{henry2019leveragingbfloat16artificialintelligence}. 
As well as E4M3 and E5M2 for $\mathrm{FP}8$, both are standard formats that are already supported by several mainstream deep learning inference engines, such as MLC-LLM, Quanto, and so on (see Section~\ref{sec2:sec:bitwidth} for details). 


\label{sec1:para:nf}
\textbf{NormalFloat (NF)}~\citep{dettmers2024qlora} is a fixed floating-point method used in weight-only quantization strategies for LLMs. The data representing format follows the floating-points, but the $2^k$ values $X^\mathrm{NF}_{i}, i\in[0, 2^k-1]$ are estimated to be:
\begin{align}
X^\mathrm{NF}_{i}
  &= \frac{1}{2}\Bigl(
      quantile\!\left(N(0,1), \frac{i}{2^{k}+1}\right) \notag\\
  &\quad +\, quantile\!\left(N(0,1), \frac{i+1}{2^{k}+1}\right)
    \Bigr),
\label{Xeqn2}
\end{align}
where 
$quantile(\cdot, q)$ is the $q$-th quantiles of the input. $N(0,1)$ means the standard normal distribution. For a tensor that does not fall within the range of -1 to 1, we must first scale it using its maximum absolute value. To ensure the exact representation for zero, it asymmetrically divides the data into the positive and negative parts by estimating $2^{k-1}$ of $X^\mathrm{NF}_{i}$ for the negative and $2^{k-1}-1$ for the positive, then removes one of the zeros in both sets. NF is estimated to have an almost equal expected number of values in each quantization bin to keep the most information in the quantized formats. 

\textbf{Micro Scaling FP} \citep{rouhani2023microscalingdataformatsdeep}. 
It was proposed and developed in collaboration with industry alliance members, including AMD, Arm, Intel, Meta, Microsoft, NVIDIA, and Qualcomm, 
which aims to establish a unified standard for fine-grained sub-blocks of tensor format. It applies E8M0 scaling factors on a block of data with various original formats (i.e., FP8, FP6, FP4, INT8). The scaling block size indicates the number of elements that each scaling applies. It keeps high precision for the value representation but is significantly efficient on hardware due to the shared scalings. 

\textbf{Integer Numbers}. Integer quantization is the most widely studied quantized data format since the quantization technique has emerged. It divides the floating-points into $2^k$ equally spaced discrete integers. 
The formula is:
\begin{equation}
X_{\mathrm{INT}_k}=(-1)^s(d_1 2^{m}+d_2 2^{m-1}+\cdots+d_{m} 2^0),\quad x\in\mathbb{N}^+,
\end{equation} 
where $m=k-1$ and $s\in\{0, 1\}$ for signed integers. $m=k$ for unsigned integers while we regard $s=0$. Therefore, the signed integers range from $[-2^{k-1}, 2^{k-1}-1]$, and the unsigned one $[0, 2^{k}-1]$. \blue{Before the advent of LLMs, integer quantization had been applied in BERT-based language models, as demonstrated by \citet{qbert}.}

\textbf{Binarized Numbers}. Binarization is the most aggressive quantization technique, which directly abstracts the sign of value~\citep{birealnet, bibert, bi-vit}. It will lose most information, but bring significant acceleration and parameter compression in inference. The hardware takes $0,1$ for each bit originally, but developers define different logic rules and accumulation algorithms to achieve various binarized computations. 
Therefore, floating-point numbers can be binarized to $\{-1, 1\}$ or $\{0, 1\}$, depending on what value we expect the single bit to represent in our algorithms. \blue{Some studies further extended binarization to ternary quantization. Before the emergence of LLMs, works such as \citet{bai-etal-2021-binarybert,zhang-etal-2020-ternarybert,bit,liu-etal-2023-binary} explored binary or ternary quantization formats.}

Table~\ref{tab:sec1:qmaxqmin} shows the representation ranges of various standard formats. It shows that even with the same bit-width, different numerical representation formats can have significantly different value ranges. The floating-point numbers with larger $E$ have larger representation ranges but sparser points. Therefore, there is a tradeoff between finer data intervals or larger data ranges when determining data formats for a specific model and task.

\small
{\begin{table}[]
\centering
\begin{tabular}{c|cc}
\hline
Format & Max (normal) & Min (normal) \\
\hline
INT4 & 7 & -8 \\
INT8 & 127 & -128 \\
FP8 (E4M3) & 448 & -448 \\
FP8 (E5M2) & 57344 & -57344 \\
FP16 (E5M10) & 65504 & -65504 \\
BF16 (E8M7) & 3.39e38& -3.39e38\\
FP32 (E8M23) & 3.40e38& -3.40e38\\
\hline
\end{tabular}
\caption{Min and Max values for different number formats~\citep{ieee754}. }
\label{tab:sec1:qmaxqmin}
\end{table}
}

\subsubsection{Custom Formats}

For faster computation and better fitting the numerical distributions of LLMs, many studies propose custom number formats besides the standard formats described above. Here we introduce three typical customized formats. 
We omit the works before the advent of LLMs \citep{adaptivefloat} because their performance has not been validated on LLMs. 

\textbf{Floating-point Integer (Flint)}~\citep{Flint} combines the advantages of floating-point and integer representations, which is $X_{\mathrm{Flint}}=2^{{p}-bias}\times(1.\texttt{mantissa})$. We take the 4-bit Flint on float-based MAC units as an example:
\begin{eqnarray}
p =& \left \{
\begin{array}{@{}l@{}l@{}l@{}} 3 - \texttt{LZD}(b_2 b_1 b_0), & \quad b_3 = 0 \nonumber\\
4+\texttt{LZD}(b_2 b_1 b_0),                     & \quad b_{3}=1\end{array},\right. 
\quad \texttt{mantissa} \nonumber\\
&= b_2 b_1 b_0 \texttt{<<} (\texttt{LZD}(b_2 b_1 b_0) + 1),
\label{Xeqn4}
\end{eqnarray}
where the $\texttt{LZD}$ denotes the \texttt{L}eading \texttt{Z}ero \texttt{D}etector~\citep{oklobdzija1994algorithmic} which accumulates the leading zeros on the left of the bitstring, \texttt{<<} is the left shift operation, and $bias=1$ for float-based Flint4. It expands the range by integrating exponents into the integers, therefore. Compared to pure integers, Flint can represent a larger range with a limited number of bits, which better fits the distribution of LLM parameters.

\textbf{Adaptive Biased Float (Abfloat)} is first proposed in Outlier-Victim Pair Quantization (OVP)~\citep{Abfloat} to deal with outliers. The difference to Flint is that Abfloat applies a bigger $bias$ to the exponent, and left shifts $m$-bit to enlarge the $1$ before \texttt{mantissa}, making the magnitude even larger to cover the outliers. The $\mathrm{E}e\mathrm{M}m$ Abfloat value can be expressed as: 
\begin{equation}
X_{\mathrm{Abfloat}} = (-1)^{s} \times 2^{p+bias} \times (2^{m}+\texttt{mantissa}). 
\end{equation}
When $\text{bias}=0$, the range is similar to $\text{Flint}4$. With $ \text{bias}=2$ for \text{E2M1}, the range changes to $\{12, \dots, 96\}$. With $\text{bias}=3$, the range further extends to $\{24, \dots, 192\}$.
The other difference to Flint is that Abfloat is only adopted on outliers, but the normal values are stored in INT4/8 or Flint4. Both data formats require custom system support to define the behavior of the base operations (such as addition, multiplication, and so on). 


\textbf{Student Float (SF)}~\citep{studentFormat} follows the floating-point format but has specific fixed points for quantization, which is different from the above two types. SF is an improvement of NF  in Section~\ref{sec1:para:nf} and holds the view that the parameters obey Student's t-distribution $S(t;\nu)$, of which the probability density function is: 
\begin{equation}
S(t;\nu)=\frac{\Gamma\left(\frac{\nu+1}2\right)}{\sqrt{\nu\pi} \Gamma\left(\frac{\nu}2\right)}\left(1+\frac{t^2}\nu\right)^{-\frac{\nu+1}2},
\end{equation}
where $t$ and $\nu$ are the independent variable and degrees of freedom, respectively, and $\Gamma$ is generalized factorial. 
\begin{equation}
\tilde{X}^\mathrm{SF}_i = quantile\left( S(t;\nu), q_i \right), \quad 
q_i = \begin{cases}
    \omega + (\frac{1}{2} - \omega)\frac{i-1}{7} & i \in \{1, \dots, 8\} \\ 
       \frac{1}{2} + (\frac{1}{2} - \omega)\frac{i-8}{8} & i \in \{9, \dots, 16\}
      \end{cases}, 
\end{equation}
where $\omega = \frac{1}{2} (\frac{1}{32}+\frac{1}{30})$, $\{q_1, \dots, q_8\}$ and $\{q_9, \dots, q_{16}\}$ are two groups of evenly spaced quantiles. Then we normalize $\tilde{X}^\mathrm{SF}$ to $[-1, 1]$ by ${X}^\mathrm{SF}_i = \frac{\tilde{X}^\mathrm{SF}_i}{\max_i|\tilde{X}^\mathrm{SF}_i|}$. 
As $\nu$ increases, the peaks of the t-distribution become shorter and wider, and SF4 spreads out more. It converges to the standard normal distribution (NF) as $\nu \to \infty$. Same as NF, SF is used in weight-only quantization (which we introduce in Section~\ref{sec2:sec:weight-only}). Therefore, it does not need the low-level definition of base operations but requires a custom dequantization from SF to standard formats.



\subsection{Quantization Granularity}
\label{sec1:quantization_granularity}

\begin{figure*}
    \centering
    \includegraphics[width=\linewidth]{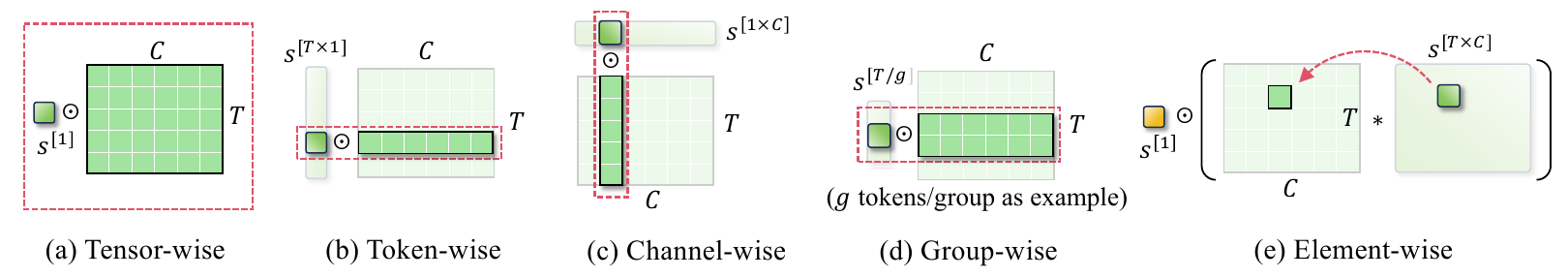}
    \caption{Illustrations for different quantization granularity. }
    \label{sec1:fig:quantization_granularity}
\end{figure*}

Quantization granularity refers to the different weight/activation partitions corresponding to each element of the scaling factor and zeropoint. It determines how finely the scale recovers and the zero point shifts. Figure~\ref{sec1:fig:quantization_granularity} showcases five fundamental types of quantization granularity: tensor-wise, token-wise, channel-wise, group-wise, and element-wise. 

\textbf{Tensor-wise} is the simplest and coarsest granularity, which takes a single scaling factor and zero point to the entire tensor~\citep{zhang2024flattenquantbreakinginferencecomputebound}. It can be the fastest but may lead to the most performance degradation because it is incapable of handling the values with a wide variation. Therefore, it is unsuitable for cases where accuracy is important or the task/model is sensitive to quantization. 

\textbf{Token-wise} is used in LLMs only, which means that each token (word or subword) has a scaling~\citep{yao2022zeroquantefficientaffordableposttraining}. It captures the fine-grained variations in different tokens. Usually, we adopt dynamic token-wise quantization for activation to reduce the quantization error and ensure diversity in generative models. 

\textbf{Channel-wise} means each channel in weight within a tensor uses one scale and can be merged into quantized weight~\citep {kim2024memory}. Token-wise activation and channel-wise weight are usually used together. Because for $i$-th token in activation and $j$-th channel in weight, the corresponding $s_{\textbf{x}_i} \in s_\textbf{x} \in \mathbb{R}^{T \times 1}$ and $s_{\textbf{w}_j} \in s_\textbf{w} \in \mathbb{R}^{1 \times C}$ can be calculated first as $s \in \mathbb{R}^{[1]}$ and multiplied to the coordinate $[i, j]$ in output matrix $\textbf{X}_O$. In this way, we preserve the generation performance with little computation overheads. 

\textbf{Group-wise} balances the computational complexity and the quantization error by grouping tensors or channels with the same scaling factor. It also reduces the storage of scaling factors by $g$ if there are $g$ tokens/channels per group~\citep{heo2023rethinking, yao2022zeroquantefficientaffordableposttraining}. 

\textbf{Element-wise} is only applied while training the weight, which is always used together with another quantization granularity, such as tensor-wise (see Figure~\ref{sec1:fig:quantization_granularity}(e)). Before inference, the element-wise scaling is merged into the quantized weight. Therefore, only the tensor-wise scale needs to be computed in inference~\citep{lee2023flexround} to recover the value magnitude. 

Different quantization granularity are always combined and adopted together. For example, \cite{lee2023flexround} uses a channel-wise scale for the Key matrix but a token-wise scale for the Value matrix based on the distribution of the data. More algorithms can be found in Section~\ref{sec4:sec:mix-precision}.   

\subsection{Dynamic and Static Quantization} 

\begin{figure*}
    \centering
    \includegraphics[width=\linewidth]{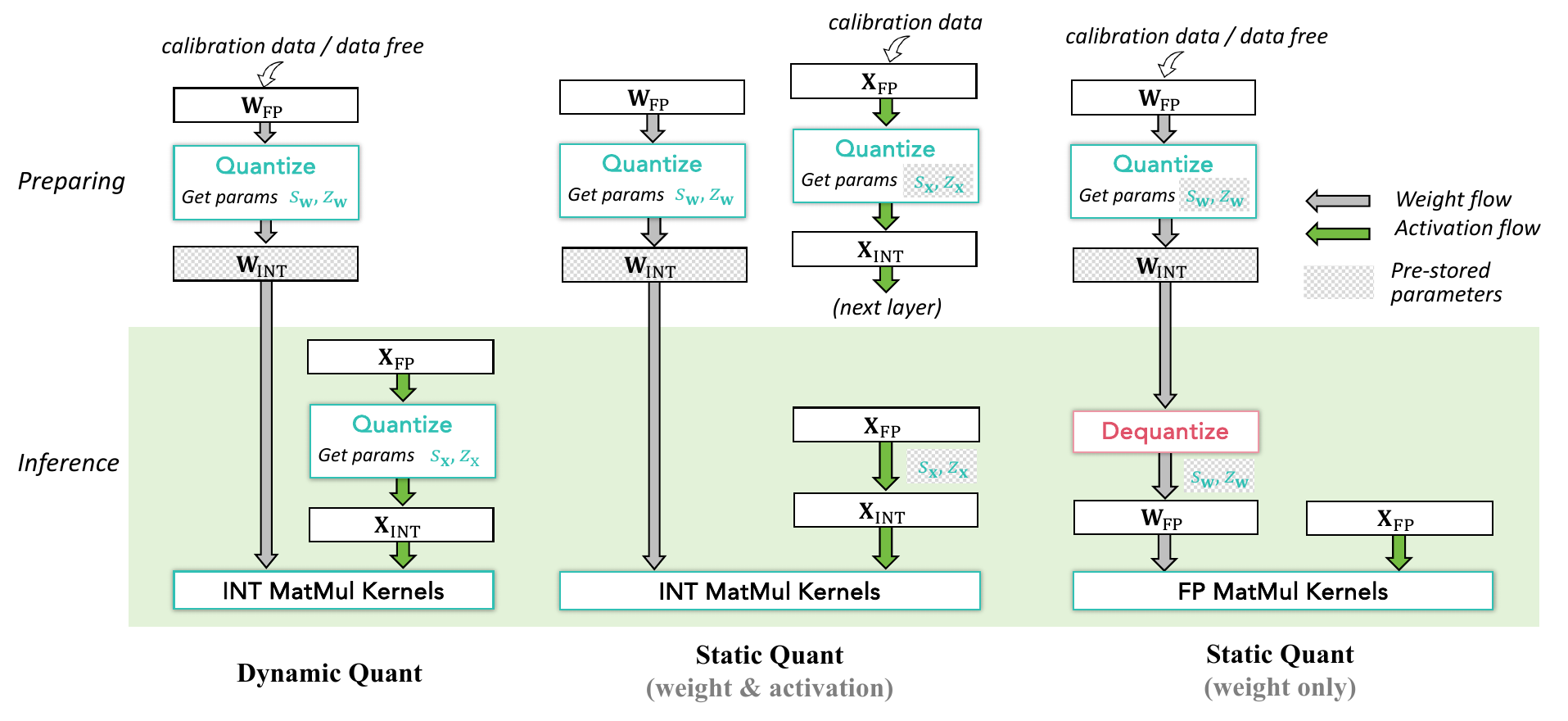}
    \caption{Dynamic and static quantization. Operations in the green block mean the inference process, while outside the block is the production and preparation process. }
    \label{sec1:fig:dynamic_static}
\end{figure*}

Dynamic and static quantization mainly refers to the strategies in PTQ, which are illustrated in Figure~\ref{sec1:fig:dynamic_static}. We take integer quantization as an example, and other low-bit quantization methods have a similar process. 

\textbf{Dynamic Quantization}~\citep{krishnamoorthi2018quantizingdeepconvolutionalnetworks, Liu_2022_CVPR} calibrates and stores quantized weight. Usually, it does not need input data, but searches for the optimal scaling factors $s_\mathbf{w}$ and zero-points $Z_\mathbf{w}$ by minimizing the quantization error for each tensor of weight. During inference, the activation will be input into the quantization module to compute the optimal scaling factors $s_\mathbf{x}$ and $Z_\mathbf{x}$, and then quantized to INT8 by the dynamically computed factors before conducting integer GEMM with quantized weight. The scaling and zero point of activation are obtained in real time based on the current batch of input data. Therefore, the scaling factor flexibly adapts the input data distribution, bringing the smallest quantization error. While it takes extra computational complexity to get the scale during inference. It is suitable for scenarios that require rapid deployment because it does not require calibration. 

\textbf{Static Quantization}~\citep{bai2021efficientposttrainingquantizationpretrained} takes calibration data consisting of a small fraction of the training dataset. By inputting the samples into the model, we find the optimal scaling factors for both weight and activation (the middle one in Figure~\ref{sec1:fig:dynamic_static}) or weight only (the right one), and are fixed during inference. It allows for the evaluation of the quantized model during preparation, ensuring that quantization does not significantly harm the model's performance. 
As for inference, the middle one in Figure~\ref{sec1:fig:dynamic_static} quantizes the activation to low-bit and computes low-bit GEMM~\citep{dettmers2022llmint8} with quantized weight. For the right one in Figure~\ref{sec1:fig:dynamic_static}, the weight will be dequantized to floating-point numbers, and the activation will not be quantized before conducting floating-point GEMM~\citep{lin2024awq}, thus we name it weight-only quantization.

\begin{table*}[ht]
    \centering \tiny
    \resizebox{\textwidth}{!}{
    \begin{tabular}{cllllllll}
    \toprule
        \textbf{Framework} & \textbf{Algorithms} & \textbf{Model family} & \textbf{Device or Engine} & \textbf{W$_{bit}$\&A$_{bit}$} & \textbf{Weight-only$_{bit}$} & \textbf{KV Cache$_{bit}$} & \textbf{Granularity}  & \textbf{Institution} \\ \hline

        TensorRT-LLM& \makecell[l]{AWQ, GPTQ, \\SmoothQuant} & \makecell[l]{Mixture-of-Expert, (e.g. Mixtral) \\ Transformer-like (e.g. Llama), \\ Multi-modal LLMs (e.g. LLaVA) } & \makecell[l]{NVIDIA GPU} & \makecell[l]{$W_\mathrm{\blue{FP}8}A_\mathrm{\blue{FP}8}$, \\ $W_\mathrm{\blue{INT8}}A_\mathrm{\blue{INT8}}$, \\ $W_\mathrm{\blue{INT4}}A_\mathrm{\blue{FP8}}$} & \makecell[l]{\blue{INT}4, \blue{INT}8} & \makecell[l]{\blue{FP}16, \blue{FP}8, \\\blue{INT}8} & \makecell[l]{tensor-wise, \\ token-wise, \\ channel-wise, \\ group-wise(128)} & \makecell[l]{NVIDIA} \\ \hline
        
        Transformers & \makecell[l]{AQLM, AWQ, \\ AutoGPTQ,\\ EETQ, HQQ, \\ ExLlama, \\ QAT, Quanto, \\ SmoothQuant} & \makecell[l]{Mixture-of-Expert, (e.g. Mixtral) \\ Transformer-like (e.g. Llama), \\ Multi-modal LLMs (e.g. LLaVA) } & \makecell[l]{x86\_64 CPU, \\ Apple MPS, \\ NVIDIA GPU, ARM CPU} & \makecell[l]{$W_\mathrm{\blue{FP}8}A_\mathrm{\blue{FP}8}$, \\ $W_\mathrm{\blue{INT}8}A_\mathrm{\blue{INT}8}$} & \makecell[l]{\blue{INT}4, \blue{INT}8} & \makecell[l]{\blue{FP}16, \blue{INT}2, \\\blue{INT}4, \blue{INT}8} & \makecell[l]{tensor-wise, \\ token-wise, \\channel-wise, \\ group-wise(128)} & {HuggingFace} \\ \hline

        OpenVINO & \makecell[l]{AWQ, GPTQ, \\ SmoothQuant} & \makecell[l]{Mixture-of-Expert, (e.g. Mixtral) \\ Transformer-like (e.g. Llama) , \\ Multi-modal LLMs (e.g. LLaVA)} & \makecell[l]{x86\_64 CPU, Intel GPU, \\ NVIDIA GPU} & \makecell[l]{ $W_\mathrm{\blue{INT}8}A_\mathrm{\blue{INT}8}$}  & \makecell[l]{\blue{INT}4, \blue{INT}8} & \makecell[l]{\blue{FP}16, \blue{FP}8, \\\blue{INT}8}  & \makecell[l]{tensor-wise, \\ channel-wise, \\group-wise(32, 64, 128)}  & \makecell[l]{Intel} \\ \hline

        ONNX-Runtime & \makecell[l]{GPTQ, HQQ, \\RTN} & \makecell[l]{Mixture-of-Expert, (e.g. Mixtral) \\ Transformer-like (e.g. Llama) , \\ Multi-modal LLMs (Phi)} & \makecell[l]{ x86\_64 CPU, \\ NVIDIA GPU} & \makecell[l]{$W_\mathrm{\blue{INT}8}A_\mathrm{\blue{INT}8}$}  & \makecell[l]{\blue{INT}4}& \blue{FP}16 & \makecell[l]{tensor-wise, \\ token-wise, \\ channel-wise, \\ group-wise(32, 128)} & \makecell[l]{Microsoft} \\ \hline

        PowerInfer & \makecell[l]{-} & \makecell[l]{Mixture-of-Expert, (e.g. Mixtral) \\ Transformer-like (e.g. Llama), \\ Multi-modal LLMs (e.g. LLaVA) } & \makecell[l]{AMD GPU, NVIDIA GPU, \\ x86\_64 CPU, Apple M CPU} & {\blue{FP}16} & \makecell[l]{\blue{INT}2, \blue{INT}3, \blue{INT}4, \\ \blue{INT}5, \blue{INT}6, \blue{INT}8} &  \blue{FP}16  &  \makecell[l]{group-wise(32, 256)} & \makecell[l]{Shanghai Jiao \\ Tong University} \\ \hline

        PPLNN  & - & \makecell[l]{Mixture-of-Expert, (e.g. Mixtral) \\ Transformer-like (e.g. Llama) } & \makecell[l]{NVIDIA GPU } & \makecell[l]{ $W_\mathrm{\blue{INT}8}A_\mathrm{\blue{INT}8}$} & {\blue{INT}8} & \makecell[l]{\blue{FP}16, \blue{INT}8, \\\blue{INT}4} &  \makecell[l]{token-wise, \\ channel-wise, \\ group-wise(128)} &   \makecell[l]{OpenMMLab \\ \& SenseTime} \\ \hline

        Xorbits Inference  & \makecell[l]{AWQ, GPTQ} & \makecell[l]{Multi-modal (Yi-VL), \\ Transformer-like (e.g. Llama), \\ Mixture-of-Expert} & \makecell[l]{CPU, Apple Metal, \\ NVIDIA GPU }& {\blue{FP}16} & \makecell[l]{\blue{INT}2, \blue{INT}3, \blue{INT}4, \\ \blue{INT}5, \blue{INT}6, \blue{INT}8} & \blue{FP}16 & \makecell[l]{group-wise(32, 256)}   & 
        {Xorbits} \\ \hline
        
        bitsandbytes  & \makecell[l]{LLM.int8()} & \makecell[l]{Mixture-of-Expert, (e.g. Mixtral) \\ Transformer-like (e.g. Llama) } & \makecell[l]{AMD GPU,  NVIDIA GPU, \\ x86\_64 CPU} & \makecell[l]{ $W_\mathrm{\blue{INT}8}A_\mathrm{\blue{INT}8}$} & \makecell[l]{\blue{INT}2, \blue{INT}4, \blue{INT}8, \\ \blue{FP}8} & \blue{FP}16 &  \makecell[l]{channel-wise}   & {HuggingFace} \\ \hline
        
        DeepSpeed-MII  & \makecell[l]{FP6-LLM, \\ ZeroQuant} & \makecell[l]{Mixture-of-Expert, (e.g. Mixtral) \\ Multi-modal LLMs (e.g. LLaVA), \\ Transformer-like (e.g. Llama) } & {NVIDIA GPU} & \makecell[l]{$W_\mathrm{\blue{INT}8}A_\mathrm{\blue{INT}8}$} & {\blue{FP}6} & \blue{FP}16 &  \makecell[l]{token-wise, \\ channel-wise} & {Microsoft} \\ \hline
        
        vLLM & \makecell[l]{AQLM, AWQ, \\ GPTQ, \\ SmoothQuant, \\ SqueezeLLM} & \makecell[l]{Mixture-of-Expert, (e.g. Mixtral) \\ Multi-modal LLMs (e.g. LLaVA), \\ Transformer-like (e.g. Llama) } & \makecell[l]{AMD GPU, NVIDIA GPU,  \\ TPU, XPU, x86\_64 CPU, \\ AWS Neuron} & \makecell[l]{$W_\mathrm{\blue{FP}8}A_\mathrm{\blue{FP}8}$, \\ $W_\mathrm{\blue{INT}8}A_\mathrm{\blue{INT}8}$} & {\blue{INT}4, \blue{INT}8, \blue{FP}6} & {\blue{FP}16, \blue{FP}8} &  \makecell[l]{tensor-wise, \\ channel-wise, \\ token-wise, \\ group-wise(32, 64, 128)} & \makecell[l]{University of \\ California, \\ Berkeley} \\ \hline

        ctransformers & \makecell[l]{ExLlama, GPTQ} & \makecell[l]{Transformer-like (e.g. Llama) } & \makecell[l]{AMD GPU, NVIDIA GPU} & {\blue{FP}16} & \makecell[l]{\blue{INT}2, \blue{INT}3, \blue{INT}4, \\ \blue{INT}5, \blue{INT}6, \blue{INT}8} & \blue{FP}16 & \makecell[l]{group-wise(32, 256)}  & \makecell[l]{Ravindra \\ Marella} \\ \hline

        MLC-LLM & - & \makecell[l]{Transformer-like (e.g. Llama) , \\ Multi-modal LLMs (e.g. LLaVA), \\ Mixture-of-Expert} & \makecell[l]{AMD GPU, NVIDIA GPU, \\ Apple Metal, Intel GPU, \\ Mobile GPU} & \makecell[l]{$W_\mathrm{\blue{FP}8}A_\mathrm{\blue{FP}8}$} & \makecell[l]{\blue{INT}3, \blue{INT}4, \blue{INT}8} & \blue{FP}16 &  \makecell[l]{tensor-wise, \\ channel-wise, \\ group-wise(32, 40, 128)}   & \makecell[l]{MLC} \\ \hline

        LMDeploy  & \makecell[l]{AWQ, GPTQ} & \makecell[l]{Mixture-of-Expert, (e.g. Mixtral) \\ Multi-modal LLMs (e.g. LLaVA), \\ Transformer-like (e.g. Llama) } & \makecell[l]{NVIDIA GPU, \\ NVIDIA Jetson} & \makecell[l]{$W_\mathrm{\blue{INT}8}A_\mathrm{\blue{INT}8}$} & \makecell[l]{\blue{INT}4}& \makecell[l]{\blue{INT}4, \blue{INT}8} &  \makecell[l]{tensor-wise, \\ token-wise, \\ channel-wise, \\ group-wise(128)}   &  \makecell[l]{Shanghai AI Lab} \\ \hline

        LightLLM & \makecell[l]{AWQ, GPTQ} & \makecell[l]{Mixture-of-Expert, (e.g. Mixtral) \\ Transformer-like (e.g. Llama) , \\ Multi-modal LLMs (e.g. LLaVA) } & \makecell[l]{NVIDIA GPU \\ NVIDIA Jetson} & \makecell[l]{ $W_\mathrm{\blue{INT}8}A_\mathrm{\blue{INT}8}$}& \makecell[l]{\blue{INT}4, \blue{INT}8, \blue{FP}6} & \makecell[l]{\blue{FP}16, \blue{INT}8}  & \makecell[l]{token-wise, \\ channel-wise, \\ group-wise(128)}  & {SenseTime} \\ \hline

        QServe & \makecell[l]{QoQ} & \makecell[l]{Mixture-of-Expert, (e.g. Mixtral) \\ Transformer-like (e.g. Llama), \\ Multi-modal LLMs (e.g. LLaVA) } & \makecell[l]{NVIDIA GPU} & \makecell[l]{$W_\mathrm{\blue{INT}4}A_\mathrm{\blue{INT}8}$, \\ $W_\mathrm{\blue{INT}8}A_\mathrm{\blue{INT}8}$} & \makecell[l]{\blue{FP}16} & \makecell[l]{\blue{FP}16, \blue{INT}8, \\\blue{INT}4} & \makecell[l]{tensor-wise, \\ token-wise, \\ channel-wise, \\ group-wise(128)}  &  \makecell[l]{MIT EECS} \\ \hline

        llama.cpp & \makecell[l]{AWQ} & \makecell[l]{Mixture-of-Expert, (e.g. Mixtral) \\ Multi-modal LLMs (e.g. LLaVA), \\ Transformer-like (e.g. Llama) , } & \makecell[l]{AMD GPU, \\ NVIDIA GPU} & {\blue{FP}16} & \makecell[l]{\blue{INT}2, \blue{INT}3, \blue{INT}4, \\ \blue{INT}5, \blue{INT}6, \blue{INT}8} & \blue{FP}16 & \makecell[l]{group-wise(32, 256)} & {ggml} \\ \hline
        
        llama2.c & - & \makecell[l]{Transformer-like (e.g. Llama) } & \makecell[l]{NVIDIA GPU, \\ Intel CPU, ARM CPU} & {\blue{FP}16} & {\blue{INT}8} & \blue{FP}16 & \makecell[l]{group-wise} & {Andrej Karpathy} \\ \hline 

        inferflow & \makecell[l]{AWQ} & \makecell[l]{Mixture-of-Expert, (e.g. Mixtral) \\ Transformer-like (e.g. Llama) } & \makecell[l]{NVIDIA GPU, \\ x86\_64 CPU, \\ ARM CPU} & \makecell[l]{\blue{FP}16 only} & \makecell[l]{\blue{INT}2, \blue{INT}3, \blue{INT}4, \\ \blue{INT}5, \blue{INT}6, \blue{INT}8} & \blue{FP}16 & \makecell[l]{group-wise(32, 256)} &  \makecell[l]{InferFlow} \\ \hline 

        ScaleLLM & \makecell[l]{-} & \makecell[l]{Mixture-of-Expert, (e.g. Mixtral) \\ Transformer-like (e.g. Llama) } & \makecell[l]{NVIDIA GPU, x86\_64 CPU} & \makecell[l]{ \blue{FP}16} & \makecell[l]{\blue{INT}4} & \blue{FP}16 &  \makecell[l]{channel-wise, \\ group-wise(32, 64, 128)}  &  {Vectorch} \\ \hline

        SGLang  &  \makecell[l]{AWQ, GPTQ}  &  \makecell[l]{Mixture-of-Expert, (e.g. Mixtral) \\ Transformer-like (e.g. Llama) , \\ Multi-modal (e.g. LLaVA)}  & NVIDIA GPU & \makecell[l]{$W_\mathrm{\blue{FP}8}A_\mathrm{\blue{FP}8}$, \\ $W_\mathrm{\blue{INT}8}A_\mathrm{\blue{INT}8}$} & {\blue{INT}4, \blue{INT}8, \blue{FP}6} & {\blue{FP}16, \blue{FP}8} &  \makecell[l]{tensor-wise, \\ channel-wise, \\ token-wise, \\ group-wise(32, 64, 128)}   & LMSYS \\  \hline
 
        \makecell{gpt-fast} & \makecell[l]{GPTQ} & \makecell[l]{Mixture-of-Expert, (e.g. Mixtral) \\ Transformer-like (e.g. Llama)} & \makecell[l]{AMD GPU, \\ NVIDIA GPU} & \blue{FP}16 &  \blue{INT}4, \blue{INT}8 & \blue{FP}16 &  \makecell[l]{channel-wise, \\group-wise}  & Pytorch    \\ \hline
        
        \makecell{FastChat} & \makecell[l]{AWQ, GPTQ} & \makecell[l]{Transformer-like (e.g. Llama)} & \makecell[l]{AMD GPU, Metal, \\ NVIDIA GPU, Intel XPU, \\ Ascend NPU } &  \blue{FP}16 & \makecell[l]{\blue{INT}4} & \blue{FP}16 & \makecell[l]{group-wise(128)} & \makecell[l]{LMSYS} \\ \hline
        
        \makecell{OpenLLM} & \makecell[l]{AWQ, GPTQ, \\SqueezeLLM} & \makecell[l]{Mixture-of-Expert, (e.g. Mixtral) \\ Transformer-like (e.g. Llama)} & \makecell[l]{AMD GPU, \\ NVIDIA GPU} &  \blue{FP}16 & \blue{INT}4  & \blue{FP}16 & \makecell[l]{group-wise(128)} &  \makecell[l]{BentoML} \\
        \bottomrule
    \end{tabular}
    }
    \caption{Inference frameworks for quantized large language models. 
    }
    \label{tab:sec2:inference_framework}
\end{table*}

\section{Frameworks and System Support}
\label{sec:system}

In the few short years since the large language model emerged, there have arisen many frameworks to support the easy usage of LLMs. 
We have selected some well-known representative frameworks and tools related to quantization, summarized and introduced them in this section according to the following categories: (1) \textbf{Inference framework for quantization}, which provides comprehensive libraries and APIs for the rapid development and deployment of LLM applications, 
(2) \textbf{System support for quantization}, which supports the underlying core functionality for quantization methods. In the following, our emphasis is on the quantization of LLMs across various frameworks and libraries. 

\subsection{Inference Framework for Quantization}

We list the representative inference frameworks in Table~\ref{tab:sec2:inference_framework}. \blue{The inference process of Large Language Models (LLMs) consists of two key stages: Prefill and Decode. During the Prefill stage, the input prompt is tokenized and processed through the model's Transformer layers to generate contextual embeddings, leveraging self-attention mechanisms to capture dependencies between tokens. This stage establishes a rich contextual representation of the input, which is stored for subsequent text generation. In the Decode stage, the model generates text autoregressively, predicting one token at a time by iteratively considering the sequence of previously generated tokens. This involves embedding lookup, attention computation, and token selection based on probability distributions. While the prefill stage processes the entire input at once, making it computationally intensive, the decode stage operates incrementally, building the output sequentially. Together, these stages enable LLMs to produce coherent and contextually relevant text, forming the foundation for optimization techniques like quantization, which aim to enhance efficiency without compromising performance.} 
Currently, no single inference framework dominates in terms of performance or usage. However, some classic deep learning frameworks, such as TensorRT-LLM\footnote{https://github.com/NVIDIA/TensorRT-LLM}, ONNX-runtime\footnote{https://github.com/microsoft/onnxruntime}, Transformers\footnote{https://huggingface.co/docs/transformers/en/index} (Huggingface), OpenVINO\footnote{https://github.com/openvinotoolkit/nncf}, PowerInfer\footnote{https://github.com/SJTU-IPADS/PowerInfer}, PPLNN\footnote{https://github.com/openppl-public/ppl.nn}, and Xorbits Inference\footnote{https://github.com/xorbitsai/inference} have integrated the support for efficient inference of large models. In addition, other inference frameworks emerged after the advent of large models that are specifically proposed for LLMs, such as bitsandbytes\footnote{https://github.com/bitsandbytes-foundation/bitsandbytes}, ctransformers\footnote{https://github.com/marella/ctransformers}, MLC-LLM\footnote{https://github.com/mlc-ai/mlc-llm}, DeepSpeed-MII\footnote{https://github.com/microsoft/DeepSpeed-MII}, vLLM\footnote{https://github.com/vllm-project/vllm}, 
LMDeploy\footnote{https://github.com/InternLM/lmdeploy}, 
LightLLM\footnote{https://github.com/ModelTC/lightllm}, QServe\footnote{https://github.com/mit-han-lab/qserve}, llama.cpp\footnote{https://github.com/ggerganov/llama.cpp}, llama2.c\footnote{https://github.com/karpathy/llama2.c}, inferflow\footnote{https://github.com/inferflow/inferflow}, ScaleLLM\footnote{https://github.com/vectorch-ai/ScaleLLM}, SGLang\footnote{https://github.com/sgl-project/sglang}, gpt-fast\footnote{https://github.com/pytorch-labs/gpt-fast}, FastChat\footnote{https://github.com/lm-sys/FastChat}, OpenLLM\footnote{https://github.com/bentoml/OpenLLM} and so on. These frameworks are lightweight and have integrated many specialized optimization techniques for large models.

\subsubsection{Ready-to-use Algorithms}
With the emergence of quantization algorithms for LLMs, some typical methods have already been integrated into most frameworks, while some methods may be developed and published originally on a specific framework. We list the most ready-to-use algorithms in each mainstream framework in Table~\ref{tab:sec2:inference_framework}. Some methods are included by most frameworks, such as GPTQ~\citep{frantar2022gptq}, AWQ~\citep{lin2024awq}, SmoothQuant~\citep{xiao2023smoothquant}, and so on. These methods share several advantages: high accuracy and efficient performance after quantization, seamless integration into existing implementation procedures, and user-friendliness. 

In addition, some algorithms are supported by several frameworks. For example, LLM.int8()~\citep{dettmers2022llmint8} was well supported by bitsandbytes (in HuggingFace), which allows to store and load 8-bit weights directly from the HuggingFace Hub and quantize weight in linear layers to 8-bit. 
FP6-LLM~\citep{xia2024fp6} is integrated in DeepSpeed-FastGen\footnote{https://github.com/microsoft/DeepSpeed/tree/master/blogs/deepspeed-fastgen} \citep{holmes2024deepspeedfastgenhighthroughputtextgeneration} to implement the runtime quantization for 6-bit floating-point weight-only quantization. It allows efficient quantization and dequantization of 6-bit weight LLMs through a unified configuration option. 
It is noteworthy that Transformers (by HuggingFace) and QServe (by MIT EECS~\cite{lin2024qserve}) integrate most algorithms with comprehensive user manuals and detailed examples, enabling a quick start for deep learning researchers and developers.

\subsubsection{Bitwidth Support}
\label{sec2:sec:bitwidth}
The support for bitwidth always reflects how comprehensive the quantization system implementation is for an inference framework or engine. It can be categorized into three types according to its position and function in accelerating LLMs: 

\textit{\textbf{Weight-only{$_{bit}$}}} means only quantizing the weight while keeping FP16 activation~\citep{lin2024awq}. The quantized weight will be dequantized back to FP16 using pre-obtained scaling factors, and then conduct FP16 \texttt{mma} with FP16 activation. Therefore, it theoretically supports non-uniform quantization with arbitrary bitwidth. The speedup is achieved by reducing the latency of data transmission between the computing device and storage host with smaller amounts of weight data, but the dequantizing of weight costs extra time. The detailed speedup will be discussed in Section~\ref{sec2:sec:weight-only}. 

\textit{\textbf{W$_{bit}$\&A$_{bit}$}} means that the algorithm quantizes both the weight and activation, and conducts low-bit matrix multiplication (MatMul) in low-level (for example, in PTX ISA 8.5\footnote{https://docs.nvidia.com/cuda/parallel-thread-execution/index.html} for NVIDIA GPUs, instruction \texttt{mma.sync.aligned.shape.row.col .s32.u4.u4.s32} means the data type of the multipliers is the 4-bit unsigned integer). All the frameworks support the INT8 and FP16 MatMul. However, limited by the computing capabilities of the hardware and the supported operations in the instruction set, only part of them have INT4 and FP8 MatMul. Few supports different bitwidth of weight and activation (like $W_\mathrm{INT4}A_\mathrm{INT8}$), which requires customized computation kernels with assembled GEMV instructions\footnote{https://huggingface.co/docs/transformers/main/en/quantization/eetq} ~\citep{egiazarian2024extremecompressionlargelanguage}. It should be mentioned that if you want to use low-bit MatMul, your hardware architecture must support the specific low-bit computing, and it is necessary to upgrade/downgrade the driver to the corresponding version to reproduce the real low-bit computation and get the desired speedup ratio. 

\textit{\textbf{KV Cache$_{bit}$}} lists the bitwidth of Key-Value Cache. As a caching technology, memory consumption of the KV cache increases rapidly as batch size and sequence length continue to grow, potentially surpassing the model size. Therefore, quantizing the KV cache significantly reduces memory usage during model inference. There are several works devoted to quantizing the KV cache~\citep{hooper2024kvquant, yue2024wkvquant, liu2024intactkv}. Similar to weight-only algorithms, the quantized key-value pairs usually need to be dequantized to floating-point before MatMul, otherwise, the specific system support of multiplying low-bit to floating-point is required. Except for the listed bitwidth, all frameworks support the FP16 KV cache, which means directly storing the activation. 

We also list the quantization granularity. Users should refer to the manual to make sure that the quantization granularity is used for weight, activation, or KV cache. We sort out the granularity supports in each framework as a reference to help choose a suitable framework that implements the desired computation kernels. 

\subsubsection{Target Platforms} 
Numerous vendors are competing fiercely in the deep learning hardware. As one of the pioneers in the field of deep learning GPUs today, NVIDIA GPUs are supported by most frameworks. Meanwhile, vLLM, bitsandbytes, llama.cpp, ctransformers, MLC-LLM, and PowerInfer also have the support for AMD GPUs. 
For some other processing units, such as TPU, XPU, Metal, and other hardware, the system support is relatively limited. Some frameworks that are devoted to generalizing LLMs to edge devices are more likely to extend the support for those platforms, such as MLC-LLM, ONNX-Runtime, and llama.cpp. 
However, it should be noted that the frameworks with support for both low-bit quantization and hardware deployment in Table~\ref{tab:sec2:quantization_toolkits} cannot guarantee the deployment of any quantized model on each listed hardware. Users should carefully refer to the manual for guidance. However, the table we compiled may help reduce the time it takes to find a suitable framework that may meet your deployment desire.

\subsubsection{Model Family}
All the frameworks support custom model definition and seamlessly integrate external model zoos, such as HuggingFace Hub. To help users quickly get started, the frameworks provide predefined specification files for commonly used models. We can roughly classify the large models into three categories: Transformer-like LLMs (e.g., Llama, Orion, Baichuan, ChatGLM, Falcon), Mixture-of-Expert(e.g., Mixtral, Mistral, DeepSeek), Multi-modal LLMs (e.g., LLaVA). 
However, not all large models included in external model zoos can be smoothly supported, because the frameworks integrate new algorithms with a lag. Therefore, users should refer to the model zoo provided by the framework, and make sure that the target model has no additional underlying system requirements before importing a new model from the external model zoo that is beyond the supported model list.

\subsection{System Support for Quantization}
\label{sec2:system_support}
In practical implementations, it is perplexing that some quantization algorithms, although reducing the bitwidth of weight or activation, do not lead to a faster inference. Therefore, a critical question comes into mind: \textit{How does quantization actually achieve real acceleration and storage saving?} To answer this question, we must first clarify the {data transmission process} involved in model inference. 

\begin{figure*}
    \centering
    \includegraphics[width=\linewidth]{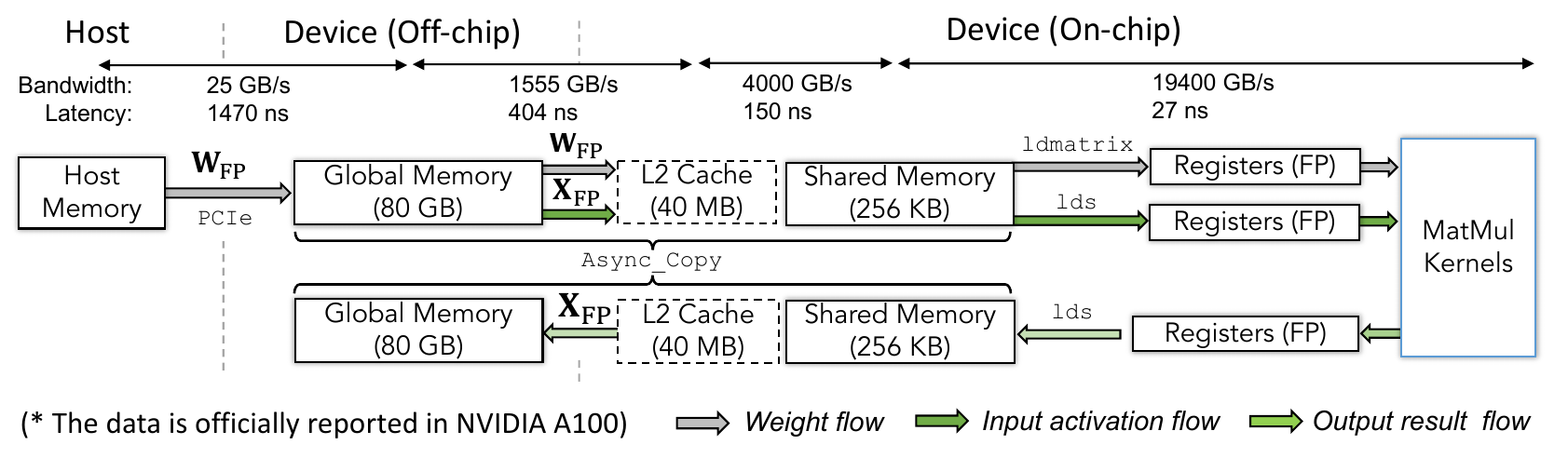}
    \caption{Data transmission of weight and activation in the caching system during inference. The bandwidth and latency are officially reported by NVIDIA A100 as an example. \blue{\texttt{PCIe} is a high-speed interface standard used for connecting various hardware components, such as GPUs, SSDs. \texttt{Async\_Copy} means asynchronous data copy using cp.async intrinsic. \texttt{ldmatrix} and \texttt{lds} are data loading instructions that load matrix from shared memory to registers with a strict layout requirement or in a fine-grained and flexible manner, respectively~\citep{NVIDIA_PTX_ISA}.}}
    \label{fig:sec2:dataflow}
\end{figure*}

\label{sec:data_transmission}
The {data transmission process of weight and activation} in the multi-level caching system is outlined in Figure~\ref{fig:sec2:dataflow}, which shows the general dataflow of quantized LLMs. GPUs typically use a hierarchical cache structure with multiple levels, each with different sizes and IO speeds. On-chip caches (L2 cache, shared memory, and registers) provide faster access but have limited capacity, while off-chip caches (device memory or global memory, host memory) offer more capacity but have slower access speed. Therefore, in today's LLMs inference frameworks, we need to load and compute data in segments with highly parallel single instruction, multiple threads (SIMT) paradigm to ensure an acceptable inference speed. 

\textit{Host memory $\rightarrow$ Device memory.} For weight, we load one layer's weight from the host memory to the device's global memory. The bandwidth is relatively low, which is 25 GB/s per direction (taking NVIDIA A100 as an example~\citep{smith2020nvidia}). If quantized, it is always in a compact format, thus the time can be saved. The activation is originally generated on the device during inference, which does not need to be copied from the host.  

\textit{Off-chip memory $\rightarrow$ On-chip memory.} We copy a chunk of weight and activation ready to compute matrix multiplication from the off-chip global memory to the on-chip L2 cache and shared memory. The amount of data copied at a time is basically determined by the design of matrix multiplication (MatMul) kernels, which is always multiple of the number of elements computed in one kernel execution by SIMT. The bandwidth is 1555 GB/s in A100. 

\textit{Shared memory $\rightarrow$ Registers.} For faster computation, the quantize/dequantize operations and MatMul are always conducted in registers. Therefore, we need to copy the weight and activation from the shared memory to the registers with small pieces. The bandwidth is 19400 GB/s, which requires more than 10 times threads and 1/780 compute intensity of \texttt{PCIe}. 

\textit{Offloading (Registers $\rightarrow$ shared memory $\rightarrow$ off-chip memory).} The computation results are copied or accumulated to the corresponding elements on shared memory. After finishing the computation for the chunk of data, the results on shared memory are offloaded to the off-chip memory. The memory that stores the weight and activation of the last chunk can be freed before moving to the next. 

Above, we have clarified the data transmission process by taking the MatMul of a linear layer as an example. 
Only after then can we answer the question: \textit{\textbf{How do quantization reduce the latency and storage?}} 
To achieve the actual inference acceleration and storage saving, we need comprehensive system support for quantization from the bottom up. 

\begin{figure*}[h]
    \centering
    \includegraphics[width=\linewidth]{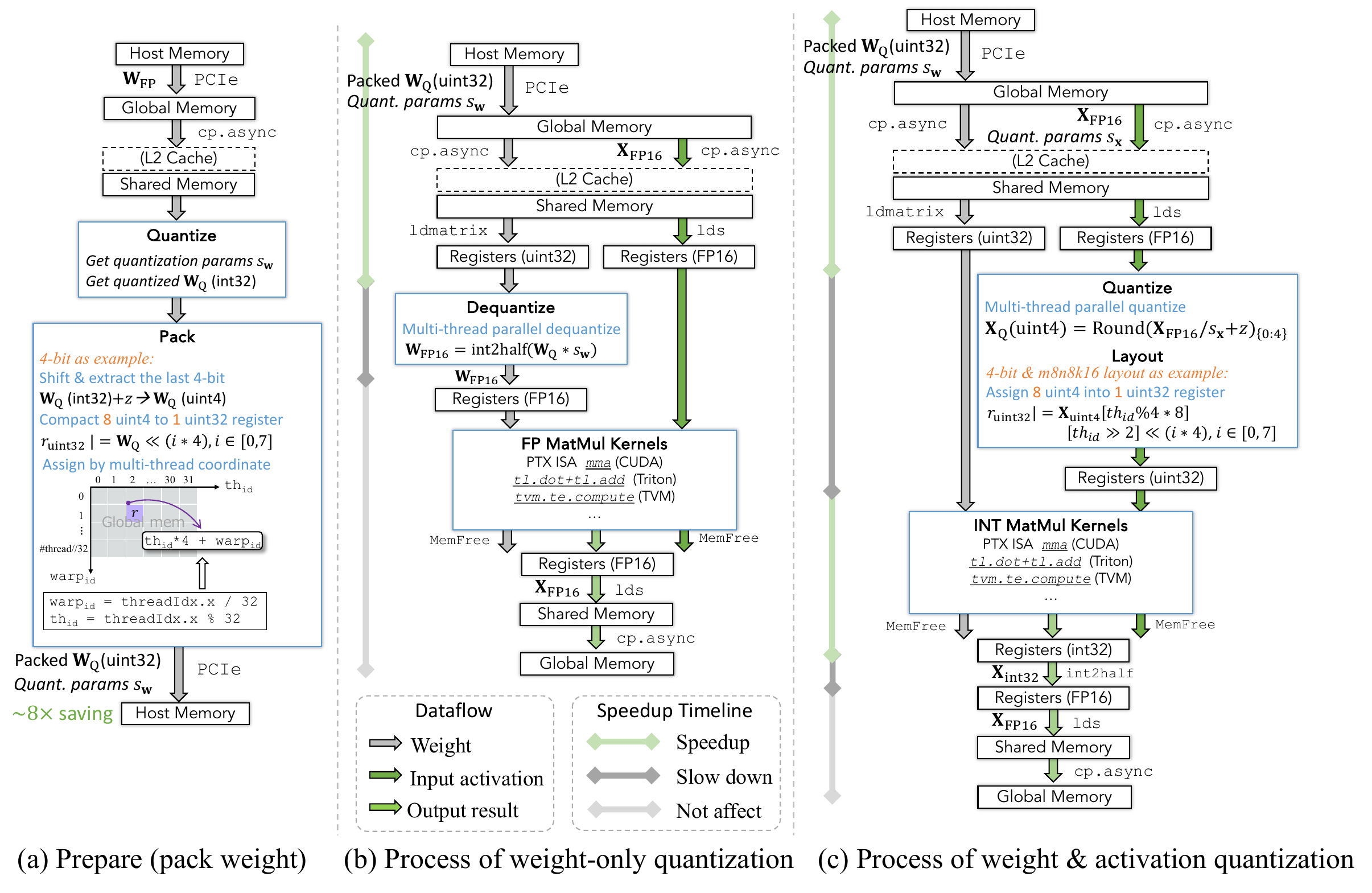}
    \caption{The data transmission process of quantization for (a) Quantized weight preparation (weight pack), (b) Weight-only quantization, and (c) Weight \& Activation quantization. }
    \label{fig:sec2:matmul_process}
\end{figure*}

In the following sections, we demonstrate the system supports for quantization according to the action scopes: \textbf{Weight-only, Weight \& Activation, KV Cache, and Quantization \& Dequantization}. 
We first provide the common and general practices in most frameworks. While these practices may not be the most efficient, they offer high scalability and generalization, allowing new algorithms and implementations to be quickly and easily integrated. Then, we introduce several custom designs. These studies investigate the speedup and generation quality bottlenecks and propose faster solutions for a certain scope. 
Figure~\ref{fig:sec2:matmul_process} shows how the quantization of weight or activation reduces inference time (4-bit integer quantization is taken as an example, which can also be any other low-bit data format). Figure~\ref{fig:sec2:kv_cache} illustrates how quantized KV Cache affects the inference.  
Speedup Timelines in both figures clearly divide the whole process into three types based on the time consumption compared to the FP16 counterpart: \textit{Speedup} (green line), \textit{Slow down} (dark grey line), and \textit{Not affect} (light grey line).

\subsubsection{Weight-Only Quantization}
\label{sec2:sec:weight-only}


The fundamental bottleneck in model inference before and after the advent of large models is the data transmission and storage costs, which are always neglected in ordinary small models. Due to the large amount of data, the transmission latency can not be overlooked, which even surpasses the computation latency and becomes the major challenge in LLM inference. Therefore, weight-only quantization emerges, which compacts the weight and reduces the data copy burden among levels of caches~\citep{lin2024awq, frantar2022gptq}. 

The processes related to weight-only quantization are illustrated in Figure~\ref{fig:sec2:matmul_process} (a) and (b). Both weight-only and weight \& activation quantization require packing weight to lower bitwidth beforehand. The weight packing is only conducted once before inference, and it costs little computation resources and time. The weight data are distributed to multi-threads, with each thread tiles a chunk of data according to the following steps: (1) quantizing the weight to lower bitwidth by pre-obtained scaling factors, (2) densely packing them into \texttt{uINT32} units without idle bits, (3) offloading and storing into host memory. Therefore, the packed weight has a significant reduction in storage compared to the floating-point one. 

See speedup timeline in (b), weight-only quantization alleviates the burden of data transmission from host memory to on-chip memory by reducing the data amounts. However, it introduces additional dequantization of weight before conducting the MatMul because the general kernels only receive the same datatype of inputs. 
As long as the time spent on dequantization is shorter than the time saved on data transmission, the weight-only quantization brings benefits in acceleration, which indeed is the case. It is the overload of parameter transmission in LLMs that makes weight-only quantization valuable in practice. Therefore, even using floating-point MatMul kernels, weight-only quantization can still accelerate the inference of LLMs. 

As for custom designs, since weight-only quantization dequantizes the weight back to FP16, it is possible to pack the weight with arbitrary bitwidths during quantized weight preparation. Many works propose 3-bit, 5-bit, 6-bit weight quantization~\citep{shi2024inferflowefficienthighlyconfigurable, frantar2022gptq, xia2024fp6}. 
Furthermore, since the quantized weight must be dequantized to higher bitwidths before MatMul, it is not necessary to design a linear surjection from low bitwidth numbers to real values. In other words, we can map the integers to arbitrary floating-point numbers, and adopt lookup tables for dequantization~\citep{studentFormat, dettmers2024qlora}. 
To make full use of storage and reduce the time of dequantizing weight during inference, researchers design customized backends on specific platforms to support efficient inference. For example, FP6-LLM~\citep{xia2024fp6} designs a complete GPU kernel to support faster FP6$\rightarrow$FP16 dequantization and the dense storage of weight. SpQR~\citep{dettmers2023spqr} has an efficient decoding backend based on GPUs to deal with the outliers by sparse quantization and achieves load balancing. 

\subsubsection{Weight \& Activation Quantization}
\label{sec2:sec:w_a_quantization}

Following the traditional practice of quantization, both weight and activation are quantized to low bitwidth, and the MatMul kernels are also implemented by low-bit instructions. 
We illustrate the speedup timeline in Figure~\ref{fig:sec2:matmul_process}(c) that the accelerated processes are weight transmission in the caching system as well as the low-bit MatMul. The extra operations are the quantization for activation from FP16 to low-bit integer before MatMul, and the datatype casting for the results from INT32 to FP16 after MatMul. 
Weight \& activation quantization yields greater acceleration compared to the weight-only quantization because the computationally intensive MatMul usually can be accelerated by lower bitwidth kernels, which use more efficient instructions and a better degree of parallelism. Meanwhile, it is recommended to simplify the complexity of activation quantization to minimize the time spent on runtime quantization. 
However, the actual speedup ratio highly depends on the hardware design, such as the number of floating-point and integer processing units. 

As for custom designs, there are two categories of techniques: 
(1) Faster Quantization and Dequantization (or datatype conversion). For example, QQQ~\citep{qqq} proposes faster FP16$\rightarrow$INT8 for quantizing activation, INT4$\rightarrow$INT8 for dequantizing weight, and INT32$\rightarrow$FP16 for casting the MatMul results to accelerate the data format conversion during inference. This work is based on \cite{kim2022says} which firstly introduces a faster INT4$\rightarrow$FP16 datatype conversion. Besides speeding up, other approaches turn to remove the process. Tender~\citep{lee2024tender} proposes a decomposed quantization technique to eliminate runtime dequantization/quantization during inference. 
(2) Faster MatMul Kernel. GEMV can be more flexible and efficient in fitting various bitwidths than GEMM, and even receives input matrices with two bitwidths, such as INT1*INT8 and INT3*INT8~\citep{wang2023bitnet}. By assembling several products of a matrix and a vector, we can get the desired results without padding or idle bits. For example, EETQ\footnote{https://github.com/NetEase-FuXi/EETQ} introduces GEMV operators which are 13-27\% faster than GEMM kernel. SqueezeLLM~\citep{kim2023squeezellm} proposes LUT-based MatMul by GEMV, which supports highly efficient 4-bit MatMul kernel on hardware architectures that do not support integer MatMul instruction. AQLM~\citep{egiazarian2024extremecompressionlargelanguage} designs W1A16 and W2A8 MatMul kernels to receive input matrices with extremely low bitwidth and calculate them directly without dequantizing or datatype conversion. 


\begin{figure*}
    \centering
    \includegraphics[width=0.90\linewidth]{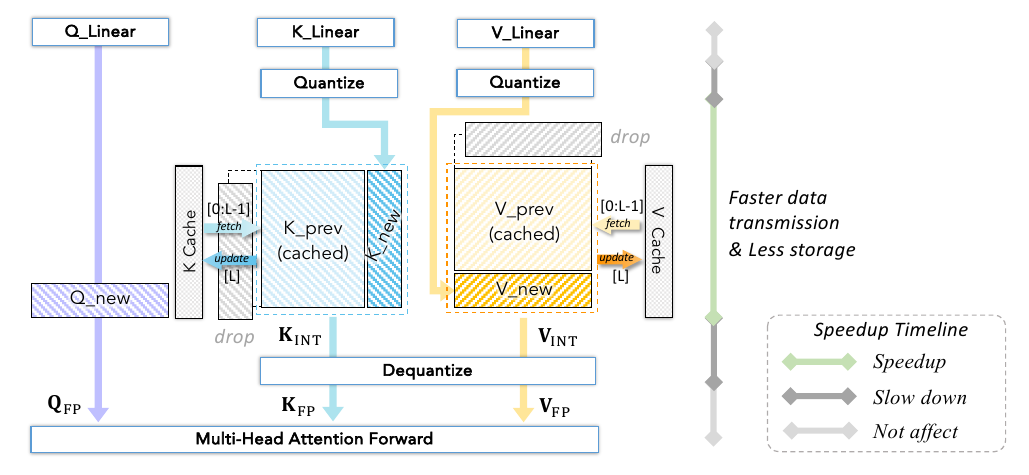}
    \caption{Illustration of KV Cache quantization. }
    \label{fig:sec2:kv_cache}
\end{figure*}

\subsubsection{KV Cache Quantization}
\label{sec2:sec:kv_cache}

KV Cache, or key-value cache, is to optimize the generative models that predict text token by token. Although the model generates only one token at a time, each token depends on the previous context. To avoid repeated calculation, the KV cache acts as a memory bank storing previous key-value results to reuse in the following generations. However, the storage highly depends on the sequence length, hidden size, attention head numbers, and so on. Quantization is an efficient approach to compressing storage. The overall process is illustrated in Figure~\ref{fig:sec2:kv_cache}. 

The KV cache is generated and updated in runtime along with the serialized input data. During inference, the $\mathbf{K}_{new}$ and $\mathbf{V}_{new}$ from linear layers are first quantized, then concatenated to the end of the stored key and value lists, which are also quantized, to form new lists. When the cache size exceeds its limit, the earliest key-value pairs will be dropped. Then we dequantize the matrices to FP16 before conducting multi-head attention forward propagation with the newly generated query $\mathbf{Q}_{new}$. We illustrate how KV cache quantization affects the inference in the Speedup Timeline. Compared to the FP KV cache, the quantized one occupies less storage in device memory and spares less time in KV data transmission in the caching system due to the smaller data bytes. 

There are mainly \blue{four} techniques for quantizing KV cache: (1) Quantizing to lower bitwidth. QoQ~\citep{lin2024qserve} compresses KV to 4-bit and proposes SmoothAttention to prevent the accuracy drop due to the lower bitwidth.  KIVI~\citep{kivi2023plug} even developed a tuning-free \blue{2-bit} KV cache quantization algorithm. \cite{yang2024no} proposes a mixed-precision strategy that quantizes the earliest KV to lower bitwidth, while keeping the new KV with more bits. (2) Quantizing window. Many studies\blue{~\citep{zhang2024kv,duanmu2024skvq}} postpone the quantization of KV pairs, but only quantize them in a batch when the length of the full-precision KV list exceeds the window size. \blue{For example, SKVQ~\citep{duanmu2024skvq} employs a sliding-window mechanism, determining the quantization parameters within the window.}
(3) Skipping the dequantization of $\mathbf{K}_{new}$. Methods like WKVQuant~\citep{yue2024wkvquant} concatenates the FP $\mathbf{K}_{new}$ and $\mathbf{V}_{new}$ to the dequantized $\mathbf{K}_{prev}$ and $\mathbf{V}_{prev}$, which preserves more information of current token in $\mathbf{K}$ and $\mathbf{V}$ matrices, then quantizes the $\mathbf{K}_{new}$ and $\mathbf{V}_{new}$ and stores them into the KV cache (when meets the condition).  
(4) Optimizing outliers. There are token-wise outliers in KV matrices, so methods such as storing the outliers with higher bitwidth or mitigating the outlier magnitudes can improve the performance~\citep{dong2024qaqqualityadaptivequantization,liu2024intactkv,kang2024gear, lin2024qserve}. We omit the details of this category, as the general practices are similar to the quantization methods used for the entire model. 

\subsubsection{Quantization and Dequantization}
\label{sec2:sec:quant_and_dequant}

In this section, we roughly categorize the quantization into three types: (1) \textit{Floating-point Quantization}, casting the high-bit floating-points into low-bit ones. (2) \textit{Integer Quantization}, which mainly refers to dividing the floating-points into evenly spaced integers. We omit requantizing higher bitwidth integers to lower bitwidth ones here, because it is seldom used in real practices, and few studies propose faster implementations to convert integers. (3) \textit{Binarization}, including \texttt{sign} and \texttt{bool} functions. 

\paragraph{\textbf{Floating-point Quantization}}

Quantizing higher bitwidth floating-point to lower is actually the clip of mantissa bits. That is because the source value with higher bitwidth usually has more or equal bits for both exponent and mantissa parts compared to the target value with lower bitwidth. Algorithm~\ref{sec1:algo_fp8_quantize} provides an example of quantizing FP32 to FP8. And we follow \cite{micikevicius2022fp8} to summarize the general process as follows: 

(1) \textit{Scale.} Since the target value occupies less bitwidth, the representation range may shrink drastically, and not be able to convey most of the data. Scaling the source value to a suitable range can best preserve the information after quantized to FP8. The scaling is pre-obtained by learning or calibration. 

(2) \textit{Check Overflow/Underflow.} Check whether the source value overflows the FP8 range, either from the upper or lower bound. If so, return the maximum or minimum directly. 
If it is not overflow, check if the exponent part {underflows} from the smallest positive normal number that the FP8 format can present. If so, we divide the value by the smallest subnormal number in FP$x$, round to the nearest integer, and then multiply the smallest subnormal number. The integer determines the value of mantissa bits and the exponent bits are all set to zero. 

(3) \textit{Copy and Round.} If the value is {neither overflowing nor underflowing} of FP8, we copy the lower $e$ bits from the source FP32 value to the target FP8 value. Then we clip the mantissa to $m$ bits by rounding to the nearest. 
It is notable that rounding and overflow/underflow handling are both crucial for maintaining numerical stability and precision in real applications. However, since the reduction of mantissa bits, precision degradation is inevitable while converting to lower bitwidth. 

\begin{algorithm}
\label{sec1:algo_fp8_quantize}
\caption{Quantization to lower-bit floating-point values. }
\begin{algorithmic}[1]
\REQUIRE $X_{\mathrm{FP}32}$, $s\in \mathbb{R^+}$ , $X_{0} \in \mathbb{R}$, $e, m \in \mathbb{z^+}$, $clip^{min}$, $clip^{max}$
\ENSURE $X_{\mathrm{FP}8}$ 
\STATE $X_\mathrm{FP32}^{unscaled} = X_\mathrm{FP32} / s$ 

\STATE $e^{min} = $ \texttt{-(1 << ($e$ - 1)) + 1}

\STATE $e^{max} = $ \texttt{(1 << ($e$ - 1))}

\STATE $m = x-e-1$ \\

{\color{ForestGreen}{// Theoretical maximum of exponent part for FP8}} 
\STATE $X_{\mathrm{FP}8}^{e} = e^{max} + 2^{8-1} << 23$  \\
{\color{ForestGreen}{// Theoretical maximum of mantissa part for FP8}} 
\STATE ${X_{\mathrm{FP}8}^{m}} = $ \texttt{\textasciitilde (0x007FFFFF >> $m$) \& 0x007FFFFF}   
\STATE $X_{\mathrm{FP}8}^{theomax} = X_{\mathrm{FP}8}^{e} + X_{\mathrm{FP}8}^{m}$

{\color{ForestGreen}{// Check exponent overflow}} \\
\IF { $X_\mathrm{FP32}^{unscaled} > \mathrm{min} (clip^{max}, X_{\mathrm{FP}8}^{theomax} )$ } 
\STATE  $ X_{\mathrm{FP}8}=\mathrm{min} (clip^{max}, X_{\mathrm{FP}8}^{theomax} )$ 

\ELSIF{$X_{\mathrm{FP}8}^{theomax} < \mathrm{max} (clip^{min}, -X_{\mathrm{FP}8}^{theomax})$}
\STATE $X_{\mathrm{FP}8} = \mathrm{max} (clip^{min}, -X_{\mathrm{FP}8}^{theomax}) $

\ELSE
\STATE $X_{\mathrm{FP}8}^{sign} = X_\mathrm{FP32}^{unscaled} $ \texttt{\& 0x80000000} 
\STATE $X_{\mathrm{FP}8}^{e} = X_\mathrm{FP32}^{unscaled} $ \texttt{\& 0x7F800000} 
\STATE $X_{\mathrm{FP}8}^{m} = X_\mathrm{FP32}^{unscaled}  $ \texttt{\& 0x007FFFFF}

{\color{ForestGreen}{// Check exponent underflow}}
\IF {$ (X_\mathrm{FPx}^{e} >> 23) - 2^{x-1} < e^{min} +1 $ }
\STATE ${X_{\mathrm{FP}8}^{min}}_{.subnorm}$ = \texttt{1 / (1 << ((1 << ($e$ - 1)) + $m$ - 2)) }
\STATE $X_\mathrm{FPx} = \texttt{round2int} (X_\mathrm{FP32}^{unscaled} / {X_{\mathrm{FP}8}^{min}}_{.subnorm}) {X_{\mathrm{FP}8}^{min}}_{.subnorm} $
\ENDIF

{\color{ForestGreen}{// Round mantissa }}
\STATE $R_m = (X_{\mathrm{FP}8}^{m} \texttt{<<} m) $ \texttt{\& 0x007FFFFF + 0x3F800000}
\STATE $R_m = \texttt{round2int} (R_m - 1)$

{\color{ForestGreen}{// Process mantissa}} \\
\STATE $X_{\mathrm{FP}8}^{m} = (X_{\mathrm{FP}8}^{m} \texttt{>>} (23-m)$ + $R_m) \texttt{<<} (23-m)$
\STATE $X_\mathrm{FP8} = X_{\mathrm{FP}8}^{sign} + X_{\mathrm{FP}8}^{e} + X_{\mathrm{FP}8}^{m}$

\ENDIF
\RETURN $X_\mathrm{FP8}$
\end{algorithmic}
\end{algorithm}

\paragraph{\textbf{Floating-point Dequantization}} 
Dequantizing floating-point numbers to higher bitwidth is straightforward. In the FP format system, the bitwidth of both the exponent and mantissa bits in lower bitwidth values will not exceed that in higher bitwidth. Therefore, we can directly extract and copy the sign bit, exponent and mantissa from the original value (with fewer bitwidth) to the most significant bits in the corresponding parts of the target value (with more bitwidth). And then we conduct zero filling on the rest bits for the exponent and mantissa parts\footnote{https://github.com/pytorch/pytorch/blob/main/c10/util/Float8\_fnuz\_cvt.h}.

\paragraph{\textbf{Integer Quantization}} We first scale the floating-point numbers to the representation span of $\mathrm{INT}k$ by dividing the scaling factor $s\in \mathbb{R^+}$, and adding a zero-point $z \in \mathbb{Z}$ to shift the clamped range~\citep{wu2020integerquantizationdeeplearning}. $\mathrm{round}(\cdot)$ is the round-to-the-nearest function, and $\mathrm{clamp}(\cdot, q^{\min}, q^{\max})$ restricts values to be within the representation span of $k$-bit with $q^{\min}=-2^{k-1}, q^{\max}=2^{k-1}-1$ in symmetric quantization and $q^{\min}=0, q^{\max}=2^k-1$ in asymmetric quantization. Therefore, the overall quantization formulation can be written as:
\begin{equation}
X_{\mathrm{INT}_{k}}=\mathrm{clamp}\left(\mathrm{round}\left(\frac{X_\mathrm{FP}}{s}\right) + z,q^{\min},q^{\max}\right),
\end{equation}
where the scaling factor $s$ can be initialized as \blue{$s_0 = ({X_{\mathrm{FP}}^{\max} - X_{\mathrm{FP}}^{\min}}) /$} \blue{$({q^{\max}-q^{\min}}), $}
where $X_\mathrm{FP}^{\max}$
     and $X_\mathrm{FP}^{\min}$ are the maximum and minimum values. 

For system support, many frameworks apply the Marlin quantization\footnote{https://github.com/IST-DASLab/marlin}~\citep{frantar2024marlin} as the standard process. 
The pseudocode Algorithm~\ref{sec1:algo2:marlin_quant} outlines the steps involved in Marlin quantization, and uses 4-bit integer quantization as an example. The values are quantized and stored as unsigned integers with the desired bitwidth. Extra pre/post-shift will be conducted to get the signed values. Therefore, we first scale the $X_\mathrm{FP32}$ values by $s$ and round it to integers. Then, adding $2^{k-1}$ to shift the values to non-negative integers within the span of uINT4 (4-bit unsigned integer). We omit the detail of C++ built-in datatype casting function \texttt{float2uint}. To be understood, we explain the packing process in lines 4 to 8 by double \textbf{for} loops, which is actually implemented as lines 9 to 11. 
In 4-bit quantization, every 8 values are packed as a single uINT32, and the quantized matrix size is a quarter of the original. By using $i$\texttt{::}8, we abstract every 8 values along the dimension $C$ starting with $i$, incrementing by 8, and ending by default (till the end of dimension $C$). And then left shift the values by $4*i$ 
to place the 4-bit value to the corresponding bit range, and leave $4*i$ zeros on the right, allowing previously-stored quantized values to be preserved after \texttt{OR} operation. 

\begin{algorithm}[htb]
\caption{Marlin quantization from FP32 to INT4.  }
\label{sec1:algo2:marlin_quant}
\begin{algorithmic}[1]
\REQUIRE $X_{\mathrm{FP32}}\in \mathbb{R}^{T, C}$, $s \in \mathbb{R^+}$
\ENSURE $X_{\mathrm{uINT}4}$
\STATE $X_{\mathrm{FP32}}^{\mathrm{round}} \leftarrow \texttt{round}(X_{\mathrm{FP32}}/{scale})$

{\color{ForestGreen}{// Shift to span of uINT4 }}
\STATE $ X_{\mathrm{FP32}}^{\mathrm{clamp}}  \leftarrow \texttt{clamp}(X_{\mathrm{FP32}}^{\mathrm{round}}  + 2^3, 0, 2^4-1)$ 

\STATE $X_{\mathrm{uINT32}} \leftarrow $ \texttt{float2uint} ($X_{\mathrm{FP32}}^{\mathrm{clamp}}$)

{\color{ForestGreen}{// Pack every 8 $X_{\mathrm{uINT32}}$ to a single uINT32}}
\FOR{ $ k \leftarrow 0 $ \TO C//8 } 
    \FOR{$i \leftarrow 0$ \TO $7$}
    \STATE $X_\mathrm{INT4}[:, k]_{(4*i+3:4*i)} \leftarrow X_{uINT32} [:, i+8*k] << (4*i) $
    \ENDFOR
\ENDFOR

{\color{ForestGreen}{// Line 4$\sim$8 can also simplify as line 9$\sim$11}} \\
{\color{ForestGreen}{// \texttt{i::8} creates a sequence starts at $i$, increments by $8$, and ends by default}}
\FOR{$i \leftarrow 0$ \TO $7$}
    \STATE $X_\mathrm{uINT4}[:, :]_{(:4*i)} \leftarrow
 X_\mathrm{uINT32}[:, i\texttt{::}8] << (4*i) $  
\ENDFOR
\RETURN $X_{\mathrm{uINT4}}$
\end{algorithmic}
\end{algorithm}

There are several custom algorithms that introduce faster data type conversions. 
QQQ~\citep{qqq} designs a faster FP16 to INT8 conversion, named \texttt{FastFP16toINT8}. It starts by shifting the FP16 values to the representation span of uINT8 by adding 128. Next, adding an additional 1024 which effectively converts and places the 8 bits of uINT8 into the lower segment of the FP16 mantissa. Finally, the lower 8 bits from FP16 are extracted and applied with an \texttt{XOR} operation with \texttt{0x80} to obtain the desired INT8 format. The overall process can be further simplified to \texttt{FMA}, \texttt{PRMT}, and \texttt{XOR} operations in practice. 

\paragraph{\textbf{Integer Dequantization}} It means projecting the integers back to the real numbers by multiplying the scaling factors, which can be expressed as:
\begin{equation}
\hat{X}_{\mathrm{FP}} = s\cdot (X_{\mathrm{INT}x} - z) \approx X_\mathrm{FP}. 
\end{equation}
Therefore, in many works $s$ can also be initialized by searching from candidates to find an optimal~\citep{wei2023outlier}: 
\begin{align}
s_\mathrm{candidate} & =  \frac{i}{num_i} s_0 , 
\quad  i \in \mathbb{Z^+}, i \in (0, num_i) \\
& s.t. ~ \min || {X}_{\mathrm{FP}}  - \hat{X}_{\mathrm{FP}} ||_p. 
\end{align}
where $num_i$ means the number of candidates, which is always set as 50, 100 and so on~\citep{ptq4vit, wei2023outlier}. $s$ can also be a learnable parameter~\citep{wei2023outlier, shao2023omniquant}. The way to find a better $s$ has been widely studied before LLMs emerged~\citep{ding2024reg, qdrop, tian2024qvd}. 

For system support, we first unpack the elements according to the way we pack them, and then multiply them to the corresponding scaling factor, which can be tensor-wise, channel-wise, token-wise, and other granularities described in Section~\ref{sec1:quantization_granularity}. Custom implementations are also proposed, \texttt{SINT4toS8}~\cite{li2023speedodysseydeployablequantization} designs a faster conversion from INT4 to INT8 by multiplying by 16.




\paragraph{\textbf{Binarization}} 
It takes the \texttt{sign} or \texttt{bool} function to abstract the sign: 
\begin{equation}
X_{\texttt{sign}} = 
\begin{cases}
1, &  X_{\mathrm{FP}} \geq 0, \\
-1, &  X_{\mathrm{FP}} < 0,
\end{cases}
\quad 
X_{\texttt{bool}} = 
\begin{cases}
1, & X_{\mathrm{FP}} \geq 0, \\
0, & X_{\mathrm{FP}} < 0.
\end{cases}
\end{equation}
Using \texttt{sign} or \texttt{bool} depends on the algorithm design, i.e. what value we expect the bits to represent. For example, binarized transformers always use \texttt{bool} function on attention scores and the post-ReLU activation. While the weight and activation in linear functions take \texttt{sign} function. Since the hardware always regards the bits as 0 or 1, we can assemble instructions to achieve any desired matrix multiplication\footnote{https://github.com/yifu-ding/BGEMM-CUDA}. 
For example, on NVIDIA GPUs, the $\texttt{mma}$ instruction takes 0/1 bit matrices and regards them as 0s and 1s while conducting bitwise accumulation operation \texttt{popcount}. 
Therefore, to obtain the correct accumulation, the \texttt{popcount} function is designed with different arithmetic rules, i.e., if it substracts 1 for each 0, we can have the result of \texttt{sign} function. 
It has many accelerated implementations, such as lookup table\footnote{https://github.com/WojciechMula}, nifty popcnt~\citep{wilkes1958preparation}, hacker popcnt~\citep{warren2012hacker}, hakmem popcnt \footnote{https://en.wikipedia.org/w/index.php?title=HAKMEM\&oldid=1228234783} and so on. 

\paragraph{\textbf{Binarization Dequantization}} 
It is simply by multiplying a scaling factor $s$, i.e. $\hat{X}_\mathrm{FP} = s \cdot X_{\texttt{sign}/ \texttt{bool}}$ to preserve the magnitude of the original values. It is easy to understand that large amounts of information will be lost in binarization. Therefore, few studies are devoted to binarizing LLMs due to the sharp performance degradation. 
Due to the significant speedup and storage reduction, it is valuable to dig deeper into binarizing LLMs, but may require a new formulation beyond \texttt{sign} and \texttt{bool} functions. DB-LLM~\citep{dbllm} proposes 2-bit weight quantization by decomposing to two 1-bit weight matrices, which can be efficient in MatMul theoretically.

\section{Quantization Strategies for Efficient LLM Training}
\label{sec:training}

\subsection{Low-bit Training}
\label{sec:low-bit-training}
There are different strategies to accelerate the training of Large Language Models (LLMs) using low-bit. The common-used techniques contain BF16, FP16, FP8, and INT8 training.

\textbf{FP16 training:} Among all the data formats, BF16 training is widely used for LLMs since they are usually stable during training. However, they require hardware (e.g., A100, 4090, H100) support with Ampere or Hopper architectures. For some older hardware like Volta or Turing architectures (e.g., V100, T4), the data format is not available. In these cases, FP16 is often adopted to speed up the training, even for some small computer vision models. Since they have smaller exponent bits, FP16 faces a higher risk of encountering underflow or overflow issues. Therefore, a loss scaling strategy is proposed to preserve small or large gradient magnitudes. A detailed process is illustrated in Algorithm~\ref{algo:fp16_training}.

\begin{algorithm}
\caption{Algorithm for Weight Update with FP16 Precision}
\begin{algorithmic}[1]
    \STATE Maintain a primary copy of weights in FP32
    \WHILE{not converged}
        \STATE Make an FP16 copy of the weights
        \STATE Forward propagation (FP16 weights and activations)
        \STATE Multiply the resulting loss with the scaling factor $S$
        \STATE Backward propagation (FP16 weights, activations, and their gradients)
        \STATE Multiply the weight gradient with $1/S$
        \STATE Complete the weight update (including gradient clipping, etc.)
    \ENDWHILE
\end{algorithmic}
\label{algo:fp16_training}
\end{algorithm}

\textbf{FP8 training.} Since some hardware vendors like NVIDIA or AMD have designated new architectures supporting FP8 or FP4 formats. To achieve satisfactory acceleration with little modification, we can utilize the library Transformer Engine provided by vendors. While the dynamic range provided by the FP8 types is sufficient to store any particular activation or gradient, it is not sufficient for all of them at the same time. This makes the single loss scaling factor strategy, which worked for FP16, infeasible for FP8 training and instead requires using distinct scaling factors for each FP8 tensor. The scaling process can be formulated as:
\begin{equation}
    FP8\_MAX = maximum\_representable\_value(fp8\_format),
\end{equation}
\begin{equation}
    exp = get\_exponent(FP8\_MAX / amax),
\end{equation}
\begin{equation}
    new\_scaling\_factor = 2.0 ^ {exp}.
\end{equation}
$fp8\_format$ indicates the formats like E4M3 or E5M2. $FP8\_MAX$ is the relevant max value under that format. $amax$ is the maximal absolute value of the tensor. Then we can calculate the $new\_$ $scaling\_factor$ with $exp$. However, the calculation of $new\_scaling\_factor$ can not be online since it will introduce much more memory access. The best practice is to employ delayed scaling. This strategy chooses the scaling factor based on the maximums of absolute values seen in some number of previous iterations. This enables the full performance of FP8 computation but requires storing the history of maximums as additional parameters of the FP8 operators. \blue{Deepseek V3~\citep{deepseekai2024deepseekv3technicalreport}, one of the state-of-the-art models, introduces fine-grained block-wise FP8 quantization, enabling highly accurate FP8 training.} In Table~\ref{tab:low_bit_training_framework}, we list the prevalent frameworks and engines that support low-bit floating-point training, including the Deepspeed\footnote{https://github.com/microsoft/DeepSpeed} from Microsoft, Megatron-LM\footnote{https://github.com/NVIDIA/Megatron-LM} from NVIDIA, and UnitScaling\footnote{https://github.com/graphcore-research/unit-scaling} from GraphCore.

\begin{table*}
\centering \tiny
    \resizebox{\textwidth}{!}{
\begin{tabular}{l|l|l|l|l}
\hline
\textbf{Institution} & \textbf{Format} & \textbf{Framework} & \textbf{Engine} & \textbf{Hardware} \\ \hline
NVIDIA & BF16 & Deepspeed, Megatron-LM & AMP & Ampere, Hopper GPUs \\ \hline
NVIDIA & FP16 & Deepspeed, Megatron-LM & AMP & Ampere, Hopper, Volta, Turing  GPUs  \\ \hline
NVIDIA & FP8 & Deepspeed, Megatron-LM & Transformer Engine (TE) & Ampere, Hopper GPUs\\ \hline
Intel & FP8 & Deepspeed, Megatron-LM & Transformer Engine (TE) & Intel Gaudi 2 AI accelerator  \\ \hline
GraphCore & FP8 & PyTorch & UnitScaling & Graphcore C600 IPU-Processor PCIe Card \\ \hline
\end{tabular}
}
\caption{Systems for low-bit training.}
\label{tab:low_bit_training_framework}
\end{table*}


\textbf{INT8 training:} During training, in addition to the model's weight parameters, it is also necessary to save the gradients required by the optimizer and the backup information of the weights or gradients. 

This makes the massive parameter scale of LLMs a more pronounced memory bottleneck during fine-tuning, hindering their deployment in broader application scenarios. INT8 Training \citep{zhu2020towards} is considered a direct method to reduce the memory usage of gradients during training. However, the instability of quantization in backpropagation makes the training of LLMs more unstable and can even lead to crashes.
QST \citep{zhang2024quantized} proposes optimizing three key sources of memory usage simultaneously: model weights, optimizer states, and intermediate activations. In addition to quantizing the LLM model weights to 4 bits and introducing a separate side network that uses the LLM's hidden states for task-specific predictions, QST also uses several low-rank adapters and gradient-free downsampling modules to significantly reduce the number of trainable parameters, thereby saving memory on optimizer states. Q-GaLore \citep{zhang2024q} points out that GaLore's memory-saving strategy of projecting gradients using SVD incurs significant time costs. To address this, Q-GaLore adaptively updates the gradient subspace based on gradient convergence statistics and keeps the projection matrix in INT4 format and the weights in INT8 format, allowing Llama-7b to be trained from scratch on a single 16GB GPU.
Jetfire \citep{xi2024jetfire} features an INT8 data flow to optimize memory access and a per-block quantization method to maintain the accuracy of pretrained transformers. 4-bit Optimizer \citep{li2024memory} uses a smaller block size and proposes to utilize both row-wise and column-wise information for better quantization, and further identifies a zero point problem of quantizing the second moment, solving it with a linear quantizer.

\begin{tcolorbox}[
    colback=blue!5,  
    colframe=blue!50, 
    title={\textbf{Takeaways of \autoref{sec:low-bit-training}}}, 
    fonttitle=\bfseries,
    boxrule=1pt,
    arc=3pt,
]
\blue{
BF16 and FP16 training have become widely adopted techniques to accelerate the training process, with relatively lower accuracy risks. FP8, while effective for fine-grained quantization in specific modules like linear layers, carries higher precision risks compared to BF16/FP16. INT8, which has been explored in some research but not yet widely adopted in practice, poses the highest accuracy risks. To mitigate these risks, techniques such as dynamic scaling are often introduced to adjust and stabilize the precision during training.
}
\end{tcolorbox}

\subsection{Quantization Strategies for PEFT}
\label{sec:peft_training}
The well-pretrained LLMs possess excellent generalization and exhibit good transferability and adaptability during fine-tuning, making them potentially useful for a variety of downstream tasks. However, the massive parameter scale of LLMs creates a significant memory bottleneck during fine-tuning, hindering their broader application. Thus, the concept of parameter-efficient fine-tuning (PEFT) is introduced to address the issue of LLM fine-tuning under resource constraints \citep{ding2023parameter, han2024parameter}.

As the demand for fine-tuning LLMs arises, it has been discovered that quantization can reduce memory usage during the fine-tuning process; some improve traditional QAT training, significantly reducing the parameter load during each update, while other class of methods combines quantization with the Low-Rank Adaptation (LoRA) fine-tuning approach.

\subsubsection{Partial Parameter Fine-Tuning with Quantization}

Previous QAT methods require almost the same resources as full parameter training, making them infeasible in resource-constrained fine-tuning scenarios. Therefore, partial parameter fine-tuning strategies have been proposed. PEQA \citep{kim2024memory} follows the naive QAT training approach. But after quantizing the weights $W$, it obtains scaling factors $s_0$ and fixed-point numbers $\overline{W}_0$, then it keeps $W$ and only trains $s_0$. OWQ \citep{lee2024owq} only updates the high-precision ``weak columns" after mixed-precision quantization.

\subsubsection{Low-bit Low-Rank Adaptation}

Low-Rank Adaptation (LoRA) \citep{hu2021lora} freezes the pre-trained weights and only trains low-rank matrices. Although it reduces the trainable parameters by 10,000 times, it does not decrease the size of the pre-trained model weight itself, thus only reducing the memory requirements for fine-tuning by 3 times.

Methods like QLoRA \citep{dettmers2024qlora} utilize low-bit quantization to further reduce the memory occupation by fine-tuning the LoRA for quantized LLMs. They first quantize the pre-trained LLM to low bits using PTQ methods:
\begin{equation}
\textbf{W}_{q} \gets quant(\textbf{W}),
\end{equation}
Where $\textbf{W}$ is the weight of each layer.

Then, they freeze all weight parameters and update only the LoRA during fine-tuning, with the forward pass as follows:
\begin{equation}
\textbf{Y} = \textbf{X} \cdot dequant(\textbf{W}\blue{_{q}}) + \textbf{X} \cdot \textbf{A} \textbf{B},
\end{equation}
Where $\textbf{X}$ is the input of each layer.

In these methods, matrix \(\textbf{A}\) is typically initialized with random Gaussian values, while \(\textbf{B}\) is initialized to all zeros. This approach not only significantly reduces the memory footprint of the model's weight parameters but also ensures that the optimizer only needs to store the gradients of LoRA during fine-tuning, greatly decreasing memory usage. QLoRA \citep{dettmers2024qlora} introduces the use of Normal Float for double quantization of \(\textbf{W}\), achieving both good accuracy retention and memory savings, allowing fine-tuning of a 65B pre-trained model using a single 48GB GPU. IR-QLoRA \citep{qin2024accurate} incorporates information theory into the QLoRA paradigm, enhancing fine-tuning performance through information calibration and connection. LoRA+ \citep{hayou2024lora+} demonstrates that setting different learning rates for matrices A and B in LoRA enables efficient feature learning. QDyLoRA \citep{rajabzadeh2024qdylora} and Bayesian-LoRA \citep{meo2024bayesian} employs more flexible rank allocation within LoRA.

Moreover, some methods aim to obtain a deployable quantized and merged model after the LoRA fine-tuning. QA-LoRA \citep{xu2023qa} uses INT format to quantize \(\textbf{W}\) and adjusts \(\textbf{X} \cdot \textbf{A}^{i \times r} \textbf{B}^{r \times o}\) to \(\mathrm{mean}(\textbf{X}) \cdot \textbf{A}^{\frac{i}{L} \times r} \textbf{B}^{r \times o}\), allowing the fine-tuned \(\textbf{A} \textbf{B}\) to be losslessly merged into the INT format \(\textbf{W}_{q}\), without extra computation when deployment. L4Q \citep{jeon2024l4q}, on the other hand, maintains the dimension \(\textbf{A} \in \mathbb{R}^{i \times r}\) and directly uses the full QAT forward propagation method, simultaneously updating the quantizer parameters \(s\) and \(b\) for \(\textbf{A}, \textbf{B}\), and \(\textbf{W} + \textbf{A} \textbf{B}\). While L4Q does not reduce the memory footprint of the weights through quantization during pre-training, the optimizer still does not need to retain the gradients of the weights, resulting in a fine-tuned quantized model that can be deployed directly with higher accuracy.

Many methods have recognized that the initialization of LoRA significantly impacts the effectiveness of these quantization-based parameter-efficient fine-tuning methods. As a result, they aim to minimize \(\left \| \textbf{W}-(\textbf{W}_q+\textbf{A}\textbf{B}) \right \|_F\) before fine-tuning. LoftQ \citep{li2023loftq} and LQ-LoRA \citep{guo2023lq} both achieve this through iterative computation: \(Q_t \gets quant(\textbf{W}-\textbf{A}_{t-1}\textbf{B}_{t-1}^\top)\) and \(\textbf{A}_t, \textbf{B}_t \gets SVD(\textbf{W}-Q_t)\). LQ-LoRA also suggests incorporating calibration data, adjusting the minimization objective to \(\left \| \sqrt F\odot(\textbf{W}-(\textbf{W}_q+\textbf{A}\textbf{B})) \right \|_F^2\), where \(F\) is the Fisher information matrix for \(\textbf{W}\), and \(\odot\) represents the Hadamard product. Additionally, LQ-LoRA introduces dynamic quantization configurations to better adapt to resource constraints.

\begin{figure*}
    \centering
    \includegraphics[width=\linewidth]{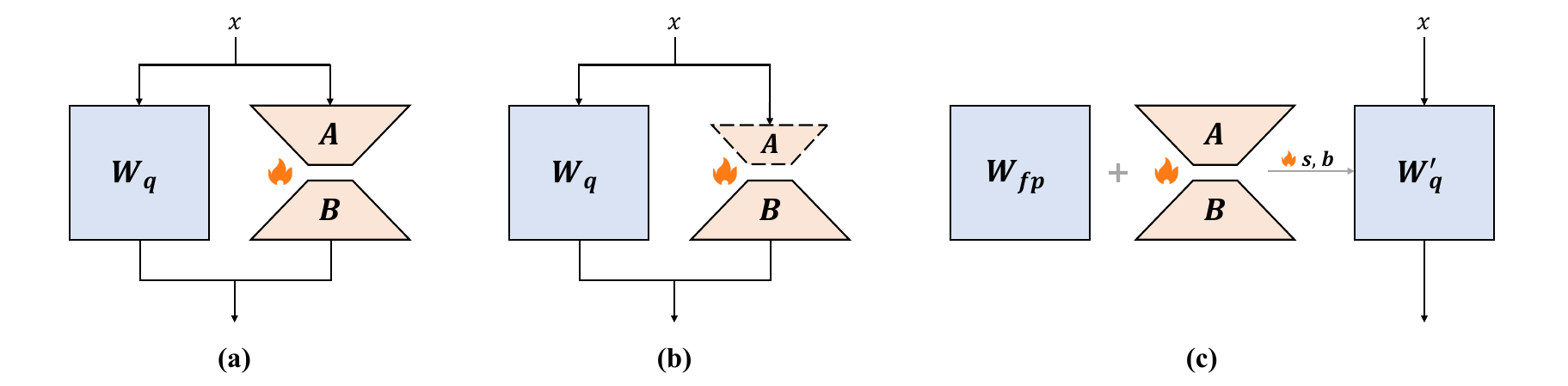}
    \caption{Illustrations for different LoRA structures.}
    \label{sec3:LoRA}
\end{figure*}

Figure~\ref{sec3:LoRA} is an illustration of different LoRA structures. Figure~\ref{sec3:LoRA}(a) represents methods like QLoRA that do not alter any part of the LLM during the fine-tuning stage and keep the complete original LoRA structure~\citep{dettmers2024qlora,qin2024accurate,hayou2024lora+,li2023loftq}. Figure~\ref{sec3:LoRA}(b) represents methods like QA-LoRA that also do not change any part of the LLM during fine-tuning but modify the original LoRA structure~\citep{xu2023qa}. Figure~\ref{sec3:LoRA}(c) represents methods like L4Q that modify the original LoRA structure and use a training process similar to QAT~\citep{jeon2024l4q}.
Both (a) and (b) require only the quantized LLM weights $W_{q}$ during fine-tuning, while (c) needs to store the pre-trained full-precision weights $W_{fp}$. (a) is solely intended to reduce training costs and cannot directly produce a quantized model after fine-tuning, while both (b) and (c) can directly integrate the LoRA module after fine-tuning to produce a deployable quantized model. \blue{Unlike the weight-only quantization in these methods, RoLoRA~\citep{huang2024rolora} incorporates rotations with LoRA for effective weight-activation quantization. Although there are existing LoRA works on MoE~\citep{li2024mixlora,luo2024moelora,gao2024higher}, they have not yet focused on quantization. In the context of quantization, it is crucial to assess whether reducing bit precision exacerbates the expert imbalance problem. Additionally, it is important to explore which position should use the LoRA-MoE method for quantization-aware training (including the router and load balancing) and to examine whether allocating more bits to deeper layers is necessary~\citep{gao2024higher}.}

\begin{tcolorbox}[
    colback=blue!5,  
    colframe=blue!50, 
    title={\textbf{Takeaways of \autoref{sec:peft_training}}}, 
    fonttitle=\bfseries,
    boxrule=1pt,
    arc=3pt,
]
\blue{
To reduce memory usage during quantization-aware training, a common strategy is to employ partial weight updates, such as updating only a subset of weight columns. To reduce memory usage during normal fine-tuning, quantization can be combined with low-rank approximation techniques, enabling fixed weights to be quantized to lower bit-widths for further memory reduction.
}
\end{tcolorbox}

\section{Quantization Algorithms for Efficient LLM Inference}
\label{sec:inference}

This section navigates through the algorithms of LLM quantization. Quantization algorithms can be broadly divided into two primary approaches: Quantization-Aware Training (QAT) and Post-Training Quantization (PTQ). QAT integrates quantization into the training/fine-tuning process, enabling the model to learn and adapt to the quantization constraints, thereby minimizing the accuracy loss associated with lower precision. In contrast, in the scenario of PTQ, we are given a pre-trained floating-point model along with a small amount of calibration data, aiming to generate an accurate quantized model without an end-to-end training process. We will delve into these quantization algorithms in detail. By the end of this section, we hope our survey can serve as a thorough and systematic collection of the various quantization algorithms applicable to LLMs, their implementation strategies, and their implications for model performance and efficiency.

\begin{table*}[]
    \centering 
    \resizebox{\textwidth}{!}{
    \begin{tabular}{llclcl}
    \toprule
         \textbf{Algorithms} & \textbf{Category} & \textbf{Target bits} & \textbf{Dataset} & \textbf{Train. time} & \textbf{Affiliation} \\ 
         \midrule
        \makecell{LLM-QAT} & \makecell[l]{W-A-KV Quant.} & \makecell[c]{4/8-bit W-A-KV} & \makecell[l]{Date-free} & medium & \makecell[l]{Meta}   \\ 
        \midrule
        \makecell{BitDistiller} & \makecell[l]{W-only Quant.} & \makecell[c]{2/3-bit W} & \makecell[l]{Alpaca, Evol-Instruct-Code,\\ WikiText-2, MetaMathQA} & fast & \makecell[l]{HKUST, SJTU, Microsoft}   \\ 
        \midrule
        \makecell{EfficientQAT} & \makecell[l]{W-only Quant.} & \makecell[c]{2/3/4-bit W} & \makecell[l]{RedPajama}& fast & \makecell[l]{OpenGVLab, HKU}   \\ 
        \midrule
        \makecell{BitNet} & \makecell[l]{W-A Quant.} & \makecell[c]{1-bit W, 8/16-bit A} & \makecell[l]{Pile, Common Crawl snapshots\\
        RealNews, CC-Stories}& slow & \makecell[l]{Microsoft, UCAS, THU}   \\ 
        \midrule
        \makecell{BitNet b1.58} & \makecell[l]{W-A Quant.} & \makecell[l]{Ternary W, 8/16-bit A} & \makecell[l]{RedPajama}& slow & \makecell[l]{Microsoft, UCAS}   \\
        \bottomrule

    \end{tabular}
    }
    \caption{Comparison of different QAT methods. }
    \label{tab:sec4:qat}
\end{table*}

\subsection{Quantization-Aware Training}
\label{sec:qat}
Table~\ref{tab:sec4:qat} summarizes the different QAT methods for LLMs. LLM-QAT~\citep{liu2023llm-qat} is the pioneering work that investigates the QAT for LLMs. To overcome the training data limits, it proposes data-free knowledge distillation which aligns the teacher logits of full-precision models and student logits of quantized models. Following LLM-QAT, BitDistiller~\citep{du2024bitdistiller} employs the asymmetric clipping strategy for asymmetric quantization during the self-distillation stage. EfficientQAT~\citep{chen2024efficientqat} significantly reduces the training cost by splitting the QAT into two consecutive phases. The first phase optimizes all parameters for each block and then the second phase merely optimizes quantization parameters for the entire network. To pave the way for a new era of extreme quantization level, BitNet~\citep{wang2023bitnet} replaces the BitLinears with original Linears and trains from scratch. Its variant, BitNet b1.58~\citep{ma2024era}, leverages ternary weight for each parameter which achieves near-lossless performance.

\begin{tcolorbox}[
    colback=blue!5,  
    colframe=blue!50, 
    title={\textbf{Takeaways of \autoref{sec:qat}}}, 
    fonttitle=\bfseries,
    boxrule=1pt,
    arc=3pt,
]
\blue{
Quantization-aware training (QAT) is particularly beneficial in extremely low-bit scenarios, despite its more complex training process. If your goal involves ultra-low-bit configurations and sufficient computational resources are available, QAT can be an effective solution. However, starting QAT from scratch can be challenging; it is generally more practical and efficient to fine-tune a pre-trained model using QAT. Additionally, it is crucial to select training data that generalizes well across diverse domains to mitigate the risk of overfitting.
}
\end{tcolorbox}

\subsection{Post-Training Quantization}
\label{sec:ptq}
\begin{figure*}[h!]
  \centering
  \includegraphics[width=\linewidth]{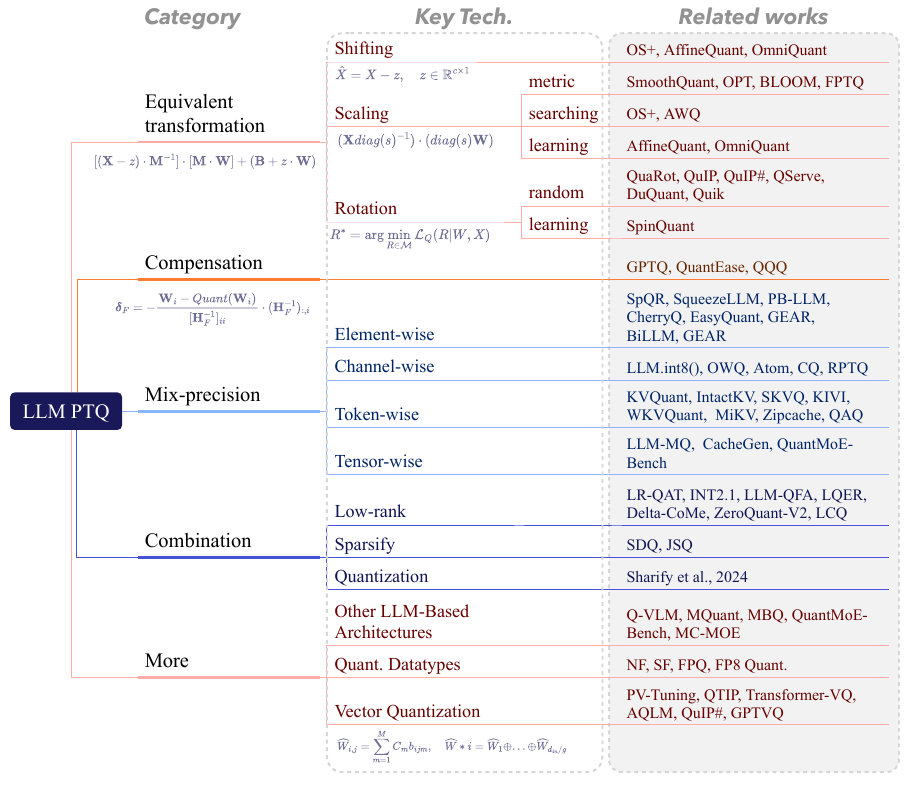 }
  \caption{An overview of the PTQ algorithms.}
  \label{fig:algo_overview}
\end{figure*}

Post-Training Quantization (PTQ) is a technique that applies quantization to a pre-trained model. Unlike QAT, PTQ does not require the model to be trained with quantization modules. This makes PTQ a highly practical approach for deploying models that were originally trained with high precision. PTQ is particularly useful when access to the training data is limited or when retraining is computationally expensive. Therefore, with the development of the LLMs, the past few years have witnessed a remarkable surge in PTQ algorithms because of their small training cost.

To have a better introduction, we systematically divide PTQ algorithms into several categories, as described in Figure~\ref{fig:algo_overview}.

\subsubsection{Equivalent Transformation}

Many studies ~\citep{luo2020positional, bondarenko2021understanding, wei2023outlier, xiao2023smoothquant} have highlighted the presence of significant outliers in LLMs. These outliers pose substantial challenges for quantization, as they force a large number of normal values to be represented with a limited number of bits, which leads to large quantization errors and accuracy degradation. Therefore, a multitude of algorithms have emerged in recent years, aiming to mitigate the issue of outliers in LLMs.

Among all the algorithms addressing the outlier problem, equivalent transformation is one of the most representative and effective methods. One of the pioneering works in applying the equivalent transformation to language models is the Outlier Suppression (OS)~\citep{wei2022outlier}. OS splits the LayerNorm function and migrates $\gamma$, which is a parameter of LayerNorm, to avoid the outlier. 
\begin{equation}
\textbf{X}_j=\textbf{X}'_j\cdot\gamma_j
\end{equation}
Then the LayerNorm becomes the non-scaling one, and the weight of the next layer can absorb the $\gamma$:
\begin{equation}
    \textbf{W}(x\odot\begin{bmatrix}
        \gamma_1 \\
        \gamma_2 \\
        \cdots \\
        \gamma_n
    \end{bmatrix})=(\textbf{W}\odot \begin{bmatrix}
        \gamma_1 & \gamma_2 & \cdots &\gamma_n \\
        \gamma_1 & \gamma_2 & \cdots &\gamma_n \\
        \cdots \\
        \gamma_1 & \gamma_2 & \cdots &\gamma_n 
    \end{bmatrix}) x
\end{equation}
By doing so, OS can suppress the outliers. Starting from the OS method, numerous subsequent equivalent transformation techniques have emerged. Most equivalent transformation methods alleviate the impact of outliers on quantization by making the outliers in weights or activations more symmetrical and smooth, which can be formulated as follows:
\begin{equation}
\label{eq:eqtrans}
    \begin{aligned}
        \textbf{Y}&=\textbf{X}\textbf{W}+\textbf{B}\\
        &=[(\textbf{X}-\Delta)\cdot\textbf{M}^{-1}]\cdot[\textbf{M}\cdot\textbf{W}]+(\textbf{B}+\Delta \cdot \textbf{W}),
    \end{aligned}
\end{equation}
where $\Delta$ is a shifting factor used to make the distribution of outliers in the input symmetrical, and $\textbf{M}$ is a matrix used to make the distribution smoother. By adopting the aforementioned equivalent transformation, many existing quantization methods have achieved state-of-the-art (SOTA) performance under various quantization settings and scenarios. 

Based on the implementation, equivalent transformation can be further subdivided into three types: shifting transformation, scaling transformation, and rotation transformation. We then independently provide a detailed introduction for each type.

\begin{figure*}[t]
  \centering
  \includegraphics[width=\linewidth]{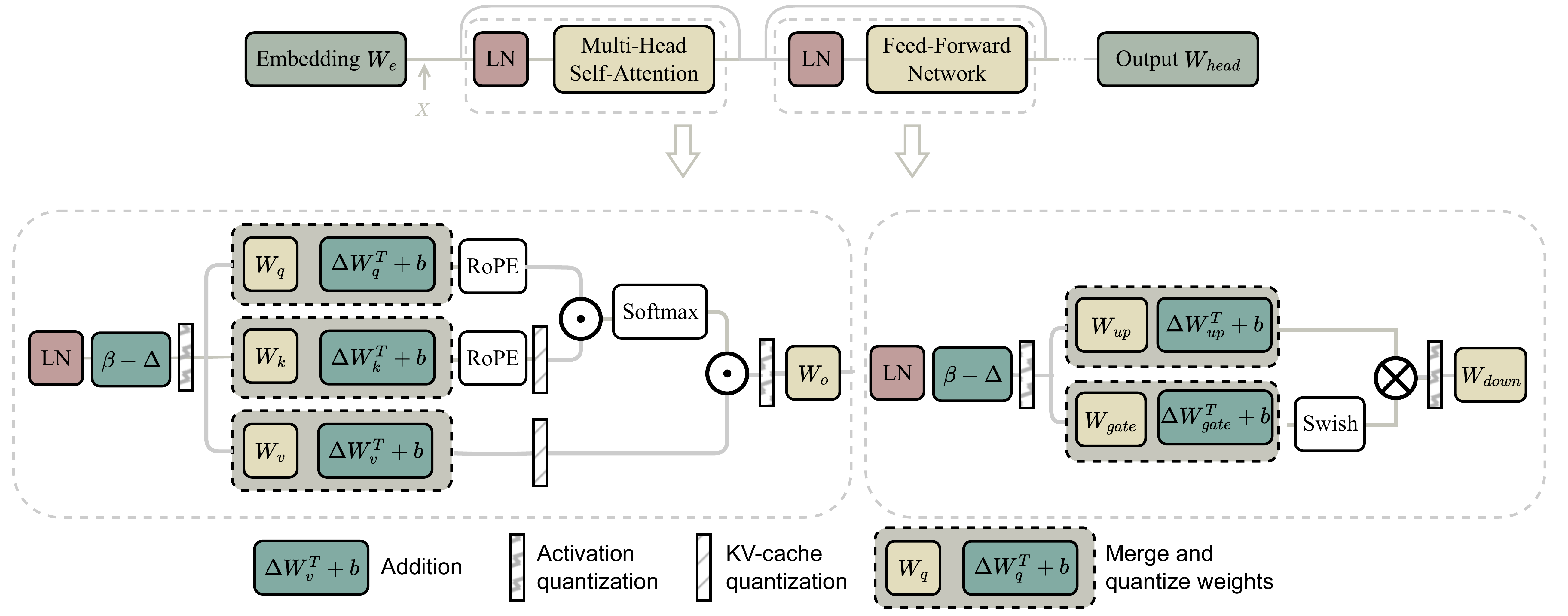 }
  \caption{Overall diagram of shifting transformation. \blue{$\Delta_1$ and $\Delta_2$ represent two types of shifting operation.} $\Delta\blue{_1}$ can be merged into the parameter $\beta$ in Layernorm and the weight metrics. \blue{Specifically, $\Delta_2$ can not be merged into the weight matrix. Therefore, the shift transformation between value projection $W_v$ and out projection $W_o$ can only be conducted online, which may raise extra computation burden.}}
  \label{fig:shifting}
\end{figure*}

\paragraph{\textbf{Shifting Transformation}} Outliers in LLMs are asymmetrically distributed across different channels. This asymmetrical representation can cause a tensor composed of channels with small ranges to exhibit a very large overall range, resulting in difficulty in the quantization process. To address this issue, OS+~\citep{wei2023outlier} first proposes the channel-wise shifting transformation, which adjusts activations across channels to mitigate the impact of asymmetry as the following equation:
\begin{equation}
    \hat{X}=X-\Delta,
\end{equation}
where $\Delta\in \mathbb{R}^{c\times1} $ serves as a row vector and shifts each channel of the activations. Note that this operation is not the conventional shifting operation used in symmetric quantization. Instead, it operates on a channel-wise basis and provides a better distribution for per-tensor quantization. In detail, OS+ defined $\Delta$ in a handicraft-way:
\begin{equation}
    \Delta_j=\frac{max(\textbf{X}_{:,j})+min(\textbf{X}_{:,j})}{2}.
\end{equation}

With the channel-wise shifting in place, the tensor range is reduced to the largest channel range, eliminating the influence of asymmetric outliers. However, handcrafting the equivalent parameters leads to sub-optimal results. Hence OmniQuant~\citep{shao2023omniquant} is proposed to determine the optimal shifting parameters in a differentiable way by including the block-wise quantization error minimization:
\begin{equation}
\label{eq:omniquant}
    \mathop{\arg\min_{\Delta}} ||\mathcal{O}(\textbf{W}, \textbf{X}) -\mathcal{O}\big(Q_w\left(\textbf{W};\Delta\right), Q_a\left(\textbf{X};\Delta\right)\big)||,
\end{equation}
where $\mathcal{O}$ represents the mapping function for a transformer block in the LLM, $Q_w(\cdot)$ and $Q_a(\cdot)$ denote the weight and activation quantizer respectively, $\Delta$ is the shifting parameter. Block-wise minimization is easy to optimize with minimal resource requirements. 
Therefore, by optimizing the objective function block by block, a more effective shifting vector can be obtained compared to the direct computation used in OS+ in an efficient and resource-saving way. However, OmniQuant requires fine-tuning of the learnable parameters; otherwise, issues such as gradient explosion can easily occur. Similar to OmniQuant, AffineQuant~\citep{ma2024affinequant} also adopts a learning-based shifting operation.

We illustrate the diagram of shifting transformation as shown in Figure~\ref{fig:shifting}. The shifting factor $\Delta$ can be fused in LayerNorm and weight matrix so that no further overhead is needed.

\paragraph{\textbf{Scaling Transformation}}

\begin{figure*}[t]
  \centering
  \includegraphics[width=\linewidth]{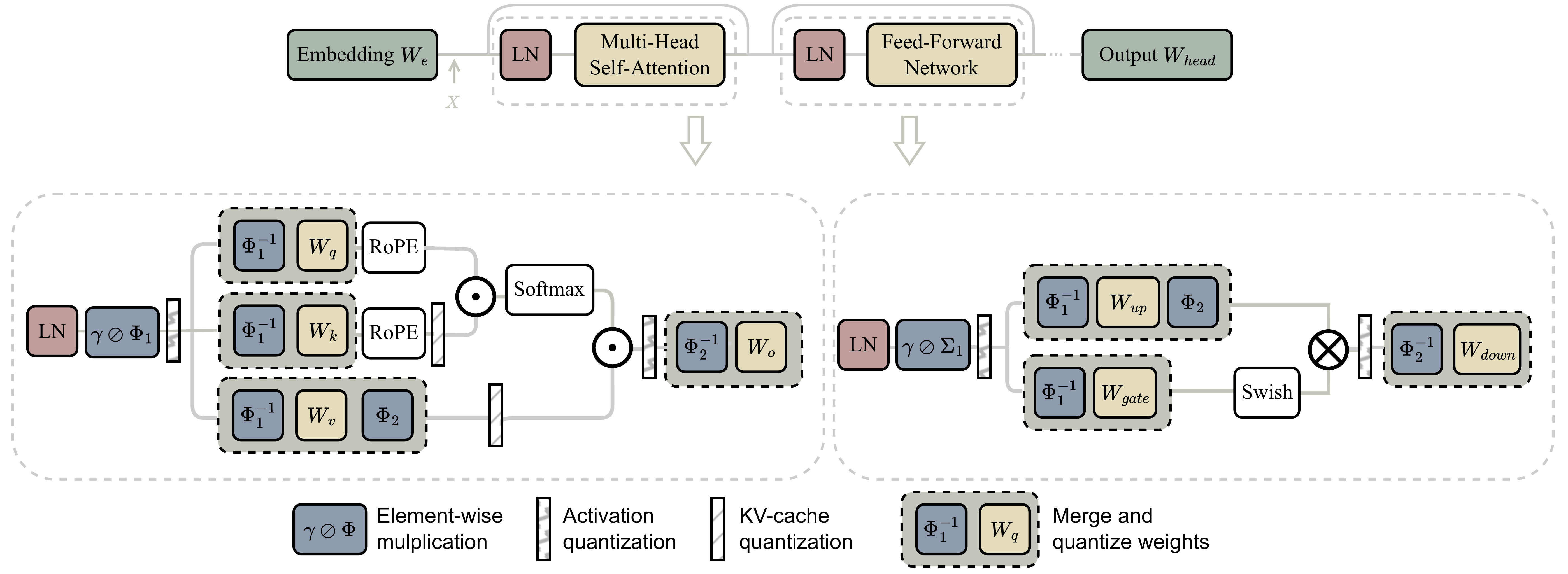 }
  \caption{Overall diagram of scaling transformation. $\Phi$ can be merged into the parameter $\gamma$ in Layernorm and the weight metrices.}
  \label{fig:scaling}
\end{figure*}

Shifting transformation effectively addresses the issue of asymmetrical distribution of outliers in activations, reducing the large range caused by the asymmetry. However, this only aids per-tensor quantization and does not reduce the difficulty of per-channel quantization, as it does not fundamentally eliminate the outliers distributed across channels in the activations. To further reduce the impact of outliers on quantization, SmoothQuant~\citep{xiao2023smoothquant} initially proposes to use a scaling transformation. It relies on a key observation: although activations are much more difficult to quantize than weights due to the presence of outliers, different tokens exhibit similar variations across their channels~\citep{dettmers2022gpt3}. Based on this observation, SmoothQuant migrates the quantization difficulty from activations to weights offline by introducing a mathematically equivalent per-channel scaling transformation that significantly smooths the magnitudes across channels:
\begin{equation}
    \textbf{Y} = (\textbf{X}diag(\Phi)^{-1}) \cdot (diag(\Phi)\textbf{W})=\hat{\textbf{X}}\hat{\textbf{W}},
\end{equation}
where $s$ is a smoothing factor. Note that $diag(\Phi)$  corresponds to the matrix $M$ in Equation~\ref{eq:eqtrans}, but it is a diagonal matrix used to achieve per-channel smoothing. SmoothQuant introduces a hyper-parameter $\alpha$ as the migration strength to control how much difficulty to migrate from activation to weights, using the following equation:
\begin{equation}
    \Phi_j = \frac{max(|\textbf{X}_j|)^\alpha }{ max(|\textbf{W}_j|)^{1-\alpha}}.
\end{equation}

However, this method requires multiple trials to determine the optimal migration strength for different models, i.e., $\alpha = 0.5$ is a well-balanced point for all OPT~\citep{zhang2022opt} and BLOOM~\citep{le2023bloom} models.

Inspired by SmoothQuant, FPTQ~\citep{li2023fptq} argues that it is unnecessary to consider weights for computing the activation smoothing scale while it is crucial to retain all the activation values with a non-linear lossless mapping. This mapping needs to fit two criteria: (1) touching gently with the inliers and (2) harshly suppressing the outliers. Based on this, they adopt a logarithmic
function to improve the calculation of the smooth matrix $s$:
\begin{equation}
    \Phi_j=\frac{max(|\textbf{X}_j|)}{ log_2(2+max(\textbf{X}_j))}.
\end{equation}

In addition to FPTQ, many other works have followed the approach of SmoothQuant. Both OS+ and AWQ~\citep{lin2024awq} use searching-based methods to find the smooth scale. However, the optimization objectives and search spaces of the two methods differ. The optimization objective of OS+ is:
\begin{eqnarray}
 \Phi^*&=&\mathop{\arg\min_\Phi}\mathbb{E}||Q\big((\textbf{X}-\Delta)\cdot diag{(\Phi)}^{-1}  \big)Q\big(diag(\Phi)\cdot\textbf{W}^\mathsf{T}\big)\nonumber\\
  &&+\hat{\textbf{b}}-(\textbf{X}\textbf{W}^\mathsf{T}+\textbf{b}) ||^2_F.
\label{Xeqn25}
\end{eqnarray}

To simplify the search space, OS+ optimizes the outlier threshold $t$, compressing channels with an activation range over $t$ into $(-t, t)$ and leaving others unchanged. This reduces the problem to a single variable. A grid search is then used for $t$ to minimize the objective. After finding the optimal $t$, the scaling vector is calculated as follows:
\begin{equation}
    \Phi_j = max(1.0, \frac{max(\textbf{X}_{:,j}-\Delta_j)}{t}).
\end{equation}

AWQ finds that the saliency of weight channels is actually determined by the activation scale. To this end, it adopts an activation-awareness optimization objective and uses a very simple search space:
\begin{align}
\Phi &= {\Phi_x}^\alpha, \notag\\
\alpha^*
  &= \mathop{\arg\min_\alpha}
     \left\| 
     Q\!\left(\textbf{W}\cdot \mathrm{diag}({\Phi_x}^\alpha)\right)
     (\mathrm{diag}({\Phi_x}^\alpha))^{-1}\textbf{X}
     -\textbf{W}\textbf{X}
     \right\|,
\label{Xeqn27}
\end{align}
where ${\Phi_x}$ is the average magnitude of activation (per channel), and use a single hyper-parameter $\alpha$ to balance between the protection of salient and non-salient channels.

In addition to searching-based methods, some approaches use learning-based techniques to find the optimal scaling matrix. OmniQuant and AffineQuant also learn the scaling matrix. In Equation~\ref{eq:omniquant}, OmniQuant learns both the shifting factor $\Delta$ and the scaling matrix $diag(\Phi)$. However, OmniQuant optimizes only within the range of a diagonal matrix. AffineQuant~\citep{ma2024affinequant} argues that this limited search range can lead to significant quantization errors, reducing the generalizability of the quantization method in low-bit scenarios. It proposes learning a general invertible matrix to perform equivalent affine transformations on weights and activations, achieving better results.

We also illustrate the diagram of scaling transformation in Figure~\ref{fig:scaling}. The same as shifting transformation, scaling factor $\Phi$ can be merged into layernorm and weight matrix.

\paragraph{\textbf{Rotation Transformation}}
Rotation transformation was first introduced by QuIP~\citep{chee2024quip}. QuIP is based on the insight that quantization works better when the weight and Hessian matrices are incoherent. This means that the weights should have similar magnitudes and the directions that require precise rounding should not align with the coordinate axes. To make it straight, a weight matrix is $\mu$-incoherent if:
\begin{equation}
    max(\textbf{W}) \leq \mu ||\textbf{W}||_F/\sqrt{mn},
\end{equation}
where $mn$ is the number of the matrix elements and $||\cdot||_F$ is the Frobenius norm. QuIP shows that multiplying a weight matrix on the left and right by an orthogonal matrix can reduce incoherence, which is equal to performing a rotation transformation on the weight matrix. QuIP utilizes Kronecker-structured orthogonal matrices, allowing for rapid additional computations. Building on this, QuIP\#~\citep{tseng2024quip} replaces these with Hadamard matrices, enhancing quantization through better incoherence and speeding up the forward pass, as the Hadamard transform can be computed in $\mathcal{O}(nlogn)$ addition operations. 

Both of these two methods target weight-only quantization. Following these approaches, QuaRot~\citep{ashkboos2024quarot} introduces a weight\&activation quantization method that also quantizes the KV cache. QuaRot operates in two stages. First, the model weights are manipulated in full precision, and two Hadamard operations are added to the model’s forward pass. In the second stage, the weights are quantized using an existing method, and quantization operations are integrated into the forward pass for online activation quantization. 

\begin{figure*}[t]
  \centering
  \includegraphics[width=\linewidth]{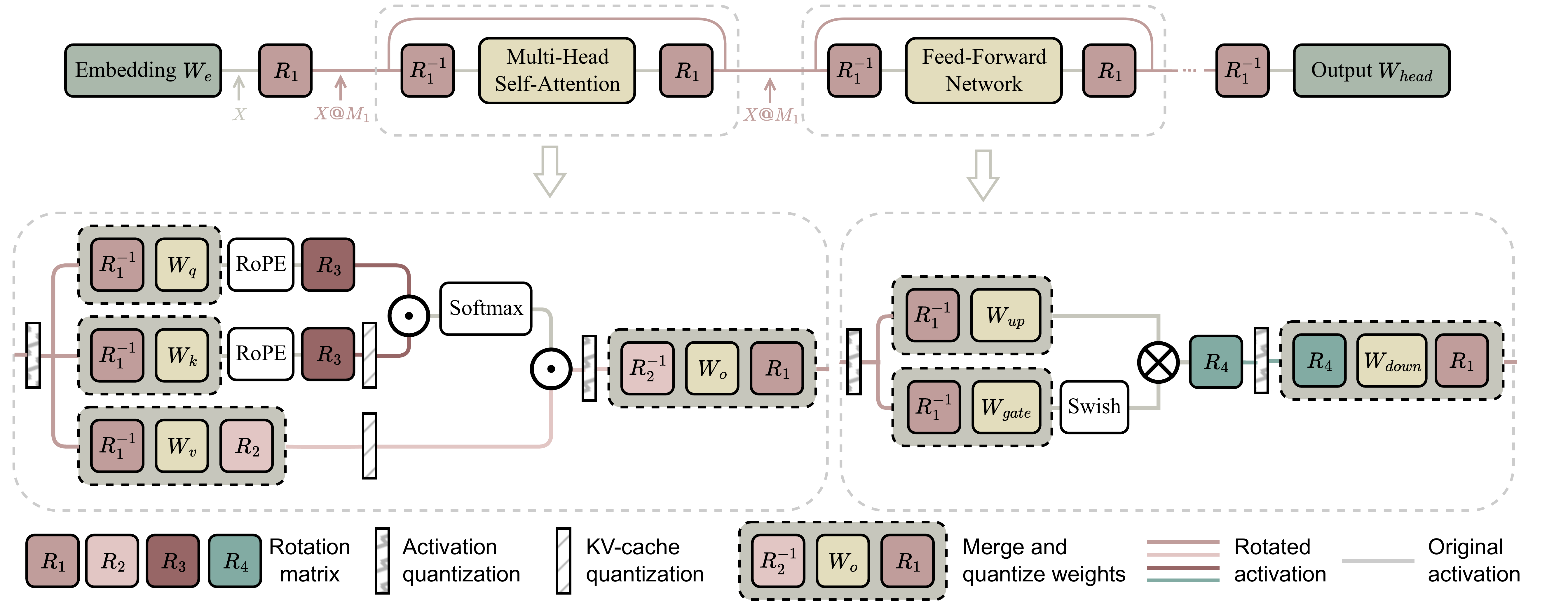 }
  \caption{Overall diagram of the rotation transformation. The rotated activations exhibit fewer outliers and are easier to quantize. $R_1$ and $R_2$ are randomized matrices which can be merged into the weights matrices. $R_3$ and $R_4$ can not be merged and are usually Hadamard matrices.}
  \label{fig:rotation}
\end{figure*}

However, both the orthogonal matrices in QuIP and the Hadamard matrices in QuIP\# and QuaRot are randomly generated. Although these works have shown that these randomly generated matrices can alleviate the outlier problem to some extent, they are not optimal.
SpinQuant~\citep{liu2024spinquant} finds that the performance of a quantized network can vary significantly with different rotation matrices. For example, the average accuracy on downstream zero-shot reasoning tasks can fluctuate by up to 13 points depending on the rotation used on the MMLU benchmark. Therefore SpinQuant proposes a learning-based rotation transformation. The rotation matrix is learned using the Cayley SGD method, with the following optimization objective:
\begin{equation}
    \textbf{R}^* = \mathop{\arg\min_{\textbf{R} \in \mathcal{M}}}\mathcal{L}_Q(\textbf{R}|\textbf{W},\textbf{X}).
\end{equation}

Here, $\mathcal{M}$ presents the Stiefel manifold, i.e., the set of all orthogonal matrices. $\mathcal{L}_Q(\cdot)$ denotes the task loss. By employing the learned matrix, the performance is improved significantly and the variance becomes much smaller compared with randomized matrices. The diagram in SpinQuant~\citep{liu2024spinquant} effectively illustrates the overall process of the rotation transformation, so we have borrowed it for our purposes as shown in Figure~\ref{fig:rotation}. \blue{Specifically, for Quarot~\citep{ashkboos2024quarot}, since it employs a head-wise rotation transformation at $R_2$, an online Hadamard matrix needs to be inserted before quantizing the attention output to achieve an equivalent transformation. DuQuant~\citep{lin2024rotation} identifies the limitations of these methods in smoothing massive outliers and therefore utilizes rotation and permutation transformations based on prior knowledge. Meanwhile, unlike SpinQuant, a greedy search strategy is employed to optimize the rotation matrix. PrefixQuant~\citep{chen2024prefixquant} discovers the token-wise outliers, especially appearing in initial tokens and low-semantic tokens. Since these tokens remain unchanged across all inputs, PrefixQuant stores their KV cache through offline prefilling.}
 
We can observe that scaling transformation and rotation transformation can be utilized for the different parts of LLM quantization. QServe~\citep{lin2024qserve} is a co-designed quantization system for efficient LLM serving, combining scaling and rotation transformations. For where additional overhead is required for the online computation of rotation matrices, QServe uses scale transformation as a substitute for rotation operations, thereby avoiding the extra overhead.

\subsubsection{Compensation}
The weight compensation technique, originally stemming from Optimal Brain Damage (OBD)~\citep{lecun1989optimal}, involves a Taylor series expansion of the objective function. This method assumes that upon the removal of any given parameter, the influence of the remaining parameters on the objective function remains unchanged. Based on OBD, OBS~\citep{hassibi1993optimal} and OBQ~\citep{frantar2022optimal} calculate the impact of each parameter weight on the objective function by solving the inverse Hessian matrix. Concurrently, they compute a compensation term applied to the remaining weights to offset the error introduced by each weight adjustment. 

Although one-by-one weight quantization methods have achieved satisfactory performance on smaller models, the computational overhead becomes prohibitive when scaling to larger models. To accelerate quantization, GPTQ~\citep{frantar2022gptq} quantizes the weights column-by-column, and the rounding errors are compensated using second-order information. Specifically, this algorithm compensates for the quantization error induced by the quantized weights $Quant(\mathbf{W}_i)$ by adjusting the subset R of full-precision weights $R$ with an update $\boldsymbol{\delta}_R$:
\begin{gather}
\mathbf{W}_i=\underset{\mathbf{W}_i}{\operatorname*{argmin}}\frac{(Quant(\mathbf{W}_i)-\mathbf{W}_i)^2}{[\mathbf{H}_R^{-1}]_{ii}}, \\
\boldsymbol{\delta}_R=-\frac{\mathbf{W}_i-Quant(\mathbf{W}_i)}{[\mathbf{H}_R^{-1}]_{ii}}\cdot(\mathbf{H}_R^{-1})_{:,i} 
\end{gather}
where the Hessian matrix is $\mathbf{H}_R = 2\mathbf{X}_R\mathbf{X}_R^\top$. Based on GPTQ, several works have been successively proposed. QuantEase~\citep{behdin2023quantease} utilizes the Coordinate Descent to compute more precise compensation for the unquantized weights. QQQ~\citep{zhang2024qqq} adopts the GPTQ for the transferred weights by OS+~\citep{wei2023outlier}. 


\subsubsection{Mixed-precision}
\label{sec4:sec:mix-precision}
As aforementioned, the presence of outliers is widely found in the activations and weights of large language models, which poses a significant challenge for quantization. Consequently, the motivation of numerous mixed-precision methods for LLMs is to represent a small number of outlier values in higher precision and other values in lower precision separately. Similarly, depending on the granularity of mixed precision, methods can be categorized into element-wise, channel-wise, \blue{token-wise} and tensor-wise approaches, as described in Section~\ref{sec1:quantization_granularity}. 

\textbf{Element-wise.} 
SpQR~\citep{dettmers2023spqr} was the first to demonstrate that outliers also exist in weights. It identifies and isolates these outlier weights based on their sensitivity, saving them as a highly sparse, higher-precision matrix. SqueezeLLM~\citep{kim2023squeezellm} adopts non-uniform quantization for non-salient weights, which achieves near-lossless performance. Similarly, CherryQ~\citep{cui2024cherry} defines heterogeneity to identify the critical cherry parameters. To explore extreme compression rate, PB-LLM~\citep{shang2023pb} is the first to binarize the non-salient weights in LLMs. Since PB-LLM still allocates high precision to 10\%-30\% of salient weights, BiLLM~\citep{huang2024billm} employs residual approximation for salient weights and group quantization for non-salient weights, reducing the quantization bit-width of LLM weights to 1.08 bits. GEAR~\citep{kang2024gear} extends the concept of mixed precision to the KV cache compression and utilizes low-rank matrices to approximate the quantization residuals.

\textbf{Channel-wise.} 
LLM.int8()~\citep{dettmers2022gpt3} splits the weights and activations into two independent parts according to the outlier channels to minimize output quantization errors in activations, which effectively reduces the GPU memory usage during inference. OWQ~\citep{lee2024owq} proposes a sensitivity-aware mixed-precision scheme to identify the weak columns by Hessian metric. Furthermore, OWQ also provides weak column tuning (WCT) to enable accurate parameter-efficient fine-tuning for task-specific adaptation. RPTQ~\citep{yuan2023rptq} observes the varying ranges across channels in activations pose challenges for quantization. Therefore, RPTQ reorders the channels into different clusters with respective quantization. Atom~\citep{zhao2024atom} employs dynamic reorder for activations and static reorder for weights to remain aligned with the corresponding activation channels. Atom further quantizes the KV cache to 4-bit which significantly boosts serving throughput. Inspired by information theory,  CQ~\citep{zhang2024kv} couples multiple key/value channels together and jointly quantize them.

\blue{\textbf{Token-wise.} Some KV cache quantization studies, such as KVQuant~\citep{hooper2024kvquant}, IntactKV~\citep{liu2024intactkv} and SKVQ~\citep{duanmu2024skvq} discover token-wise outliers caused by special tokens (first token or low-semantic-value tokens) significantly influence the performance. So they store these token-wise outliers with higher precision in advance. KIVI~\citep{liu2024kivi} and WKVQuant~\citep{yue2024wkvquant} keep the most recent KV cache in full-precision and quantize the past KV cache. MiKV~\citep{yang2024no}, Zipcache~\citep{he2024zipcache}, and Snapkv~\citep{li2024snapkv} retain the important KV pairs in high precision based on distinct metrics. QAQ~\citep{dong2024qaqqualityadaptivequantization} dynamically allocates the adaptive bits for the different tokens.}

\textbf{Tensor-wise.} 
\blue{LLM-MQ~\citep{li2023llm} assigns higher bit-widths to more sensitive layers based on first-order information and quantization error. CacheGen~\citep{liu2024cachegen} identifies LLM is more sensitive to losses in the KV cache values of the early layers than to losses in those
of the deeper layers. It assigns higher-bit precision in sensitive early layers.} The QuantMoE-Bench~\citep{li2024examining} investigates the weight bits among different blocks, experts, and linear layers, revealing that the varying numbers of weight bits are effective. 


\subsubsection{Combination}
Although current quantization methods for large models have achieved relatively good results, their performance under extremely high compression rates is still unsatisfactory due to the limited representation capacity of low-bit quantization. Currently, commonly used compression methods including low-rank decomposition, model sparsification, and model distillation are explored to combine with quantization.

\paragraph{\textbf{Low-rank}}
Although QAT is generally considered to offer the best accuracy, its high memory cost makes it difficult to apply to LLMs. Therefore, some methods consider introducing LoRA or other matrix decomposition methods as a trade-off between PTQ and QAT. Unlike PEFT discussed in Section 3.3, these methods aim to reduce quantization error using techniques like LoRA or SVD to achieve a quantized model closer to the full-precision model, rather than enhancing learning ability on fine-tuning datasets.
Some works have used LoRA to achieve parameter-efficient QAT. LR-QAT \citep{bondarenko2024low} computes $s \cdot \mathrm{clamp}(\textbf{W}_q+\textbf{A}^{i \times r} \textbf{B}^{r \times o})$ during the forward pass and does not update $\textbf{W}$ during the backward pass, allowing a 7B LLM to be trained on a single consumer-grade GPU with 24GB of memory. This approach results in a quantization-friendly model after fine-tuning. LLM-QFA aims to produce models with various bit widths through a single supernet training, significantly reducing the resource overhead of this production method by leveraging the low resource cost of LoRA. INT2.1 \citep{chai2023int2} utilizes LoRA to shift the optimization target from minimizing per-layer or per-block quantization error to minimizing the overall output error of the model. Through end-to-end fine-tuning, it reduces the distance between the output distribution and its corresponding original full-precision counterpart.
Other works have reduced quantization errors through matrix decomposition. LQER \citep{zhang2024lqer} applies SVD to quantization errors and uses an activation-induced scaling matrix to guide the singular value distribution toward the desired pattern. Delta-CoMe \citep{ping2024delta} discovers that the singular values of delta weights exhibit a long-tailed distribution after applying SVD, and proposes a mixed-precision delta quantization method that uses high-bit representations for the singular vectors corresponding to these singular values. ZeroQuant-V2 \citep{yao2023zeroquant} introduced an optimized low-rank compensation method that enhances model quality recovery by leveraging a low-rank matrix obtained through SVD of the quantization errors. LCQ \citep{cai2024lcq} uses low-rank codebooks with a rank greater than one for quantization, addressing the issue of accuracy loss when using rank-one codebooks under high compression ratios.

\paragraph{\textbf{Sparsification}}
Model sparsification aims to remove unimportant weights to accelerate the model, while quantization further reduces the remaining weights using lower-bit representations. Therefore, the two methods can be effectively used in a complementary manner. SDQ~\citep{jeong2024sdq} first sparsifies the weights of LLMs based on the magnitude as much as possible until the quality of the LLM is significantly impacted (e.g., a 1\% increase in perplexity). Then it utilizes a mixed-precision quantization method to deal with the outliers. However, this method does not take into account the coupling of the two approaches. 
Sparsification and quantization often conflict with each other. Sparsification tends to preserve parameters with large absolute values in LLMs~\citep{han2015deep, sun2023simple}, while quantization prefers a smaller range of parameter values~\citep{wei2023outlier}. As a result, the parameters preserved during sparsification may degrade the performance of quantization. JSQ~\citep{guo2024compressing} design a new sparsity metric to address this issue:
\begin{equation}
\begin{aligned}
    \textbf{I}_{ij}&=||\textbf{X}||_2\cdot||\textbf{W}||, \\
    \textbf{A}_{ij} &= max(\hat{\textbf{Y}}_{:i})-min(\hat{\textbf{Y}}_{:i}), \\
    &where\quad \hat{\textbf{Y}} = \textbf{X} \cdot (\Theta (\textbf{W};i;j))^\mathsf{T}, \\
    \textbf{S}_{ij} &= \textbf{I}_{ij} + \lambda \textbf{A}_{ij}.
\end{aligned}
\end{equation}

Here, $\Theta (\textbf{W};i;j)$ denotes an auxiliary weight matrix when setting the element at $i$th row and $j$th column as 0 in $\textbf{W}$. $\lambda$ is a trade-off factor. By using this metric, a better trade-off between preserving outliers for more information and minimizing the activation range for better quantization can be achieved.

\paragraph{\textbf{Quantization}}
In addition to combining quantization with other compression methods, different quantization techniques can also be integrated to achieve better results. A recent work~\citep{sharify2024combining} combines the SmoothQuant and GPTQ together. Actually, most of the equivalent transformation methods and the compensation quantization methods are orthogonal which can be merged for further exploration.

\subsubsection{\blue{More LLM-Based Architectures}}
\blue{Besides traditional dense LLMs, the quantization methods tailored for multimodal large language models (MLLMs) and mixture-of-expert (MoE) models have also garnered widespread attention. Q-VLM~\citep{wang2024q} offers the first post-training quantization framework for MLLMs by mining cross-layer dependency  to achieve satisfying trade-offs
between discretization errors and the search cost. MQuant~\citep{yu2025mquant} proposes a static solution, utilizing seperate quantization parameters for visual and language modality. Furthermore, it relieves weight outliers arising from online Hadamard rotations. MBQ~\citep{li2024mbq} also considers the sensitivity between language and vision modality, which adjusts the reconstruction loss for the optimal channel-wise equalization factors. QuantMoE-Bench~\citep{li2024examining} explores the structure-aware mix-precision quantization schemes for MOE models, indicating different MoE structures require varying numbers of bits. MC-MOE~\citep{huang2024mc} converts the bit allocation problem into a Linear Programming (LP) problem and balances the importance between each expert.}
\subsubsection{More Quantization Forms}
Beyond integer quantization, more forms of quantization are being introduced for LLMs, as they can also compress the average bit-width of a 32-bit or 16-bit model down to 4 or lower bits.
While these methods do not always offer significant acceleration benefits when saving memory, they generally lead to improvements in precision.

\paragraph{\textbf{More Quantization Datatypes}}
Integer quantization typically assigns a single scaling factor to an entire block and quantizes each element individually into an integer number. This reduces memory usage while also enabling the acceleration of fixed-point operations after quantizing both weights and activations. However, as higher precision is demanded for LLM quantization, formats that better match the original distribution of values have been proposed.
Normal Float~\citep{dettmers20218,dettmers2024qlora}, proposed alongside Quantile Quantization, is based on the assumption that the weight distribution follows a normal distribution. It is considered an information-theoretically optimal data type that ensures each quantization bin has an equal number of values assigned from the input tensor. However, Dotzel et al.~\citep{dotzel2024learningstudentsapplyingtdistributions} conducted a statistical analysis and found that the distributions of most LLM weights and activations follow a Student's t-distribution. Based on this, they derived a new theoretically optimal format, Student Float (SF4).
Floating-Point (FP) quantization offers better hardware support compared to NF/SF and is more flexible than integer quantization, allowing it to more effectively handle long-tail or bell-shaped distributions. Since FP can support flexible allocation of exponent and mantissa bits, several allocation schemes have been proposed. FPQ \citep{liu2023llm} determines FP quantizers through a joint format and max value search combined with a pre-shifted exponent bias. FP8 quantization \citep{kuzmin2022fp8} tests various allocation schemes by evaluating metrics like quantization error and proposes FP8 quantization simulation for learnable allocation and quantization.

\paragraph{\textbf{Vector Quantization}}
Vector Quantization (VQ) quantizes multiple vector dimensions jointly. It achieves this by learning codebooks $C_1,...,C_M$, each containing $2^B$ vectors (for B-bit codes). To encode a given database vector, VQ splits it into sub-groups of entries, and then encodes every group by choosing a vector from the learned codebook. A part of the weights of $i$-th layer is encoded by choosing a single code from each codebook and summing them up:
\begin{equation}
\widehat{W}_{i,j} = \sum_{m=1}^{M}{C_m d_{ijm}}
\end{equation}
where $d_{i j m} \in \mathbb{R}^{2^B}$ represents a one-hot code for the $i$-th output unit, $j$-th group of input dimensions and $m$-th codebook. 

To represent the full weights of $i$-th layer, simply concatenate:
\begin{equation}
\widehat{W}*{i} = \widehat{W}_{1} \oplus ... \oplus \widehat{W}_{d_{in}/g}
\end{equation}
where $\oplus$ denotes concatenation. 

Transformer-VQ~\citep{lingle2023transformer} applies vector quantization (VQ) to the key vector sequence of Attention, reducing the complexity of Attention to linear. Most other VQ works focus on optimizing the codebooks $C_m \in \mathbb{R}^{2^B}$, and the discrete codes represented by one-hot $d$. AQLM~\citep{egiazarian2024extreme} learns additive quantization of weight matrices in an input-adaptive fashion and jointly optimizes codebook parameters across each transformer block. QuIP\#~\citep{egiazarian2024extreme} uses vector quantization to exploit the spherical sub-Gaussian distribution inherent in incoherent weights by introducing a hardware-efficient codebook based on the highly symmetrical E8 lattice. GPTVQ~\citep{van2024gptvq} interleaves the quantization of one or more columns with updates to the remaining unquantized weights, using information from the Hessian of the per-layer output reconstruction MSE, and further compresses the codebooks by using integer quantization and SVD-based compression. PV-Tuning~\citep{malinovskii2024pv} notes that using straight-through estimators (STE) leads to suboptimal results and proposes an alternating iterative optimization strategy for scales, codebooks, zeros (continuous parameters), and assignments (discrete codes) during fine-tuning. QTIP~\citep{tseng2024qtip} uses a stateful decoder that separates the codebook size from the bitrate and effective dimension to achieve ultra-high-dimensional quantization.

\begin{tcolorbox}[
    colback=blue!5,  
    colframe=blue!50, 
    title={\textbf{Takeaways of \autoref{sec:ptq}}}, 
    fonttitle=\bfseries,
    boxrule=1pt,
    arc=3pt,
]
\blue{
For standard Post-Training Quantization (PTQ) of LLMs, equivalent transformation techniques such as shifting, scaling, and rotation can be employed to mitigate the impact of outliers. Quantization error can be further minimized through advanced compensation methods like GPTQ. For scenarios prioritizing high accuracy, mixed-precision quantization can be applied to recover performance loss. Conversely, if a high compression rate is the goal, combining low-rank approximation and sparsity-based methods can be effective. Furthermore, there are unique opportunities to explore emerging data formats, novel quantization functions, and cutting-edge model architectures, such as Multimodal Large Language Models (MLLMs) and Mixture of Expert (MOE) models.
}
\end{tcolorbox}

\subsection{Quantization Toolkit and Benchmark}
\label{sec:toolkit}

\subsubsection{Toolkits}  

To quantize the LLMs, there are always three basic strategies, quantization aware-training (QAT), post-training quantization (PTQ), and parameter-efficient fine-tuning (PEFT).

\begin{table*}[]
    \centering \tiny
    \resizebox{\textwidth}{!}{
    \begin{tabular}{clllll}
    \toprule
        \textbf{Toolkit} & \textbf{Algorithms} & \textbf{Model family} & \textbf{Evaluation} & \textbf{Backends} & \textbf{Institution} \\ \hline
        
        \makecell{LMQuant} & \makecell[l]{AWQ, Atom, GPTQ, QoQ, \\ QuaRot, SmoothQuant} & \makecell[l]{Mixture-of-Expert, (e.g. Mixtral), \\ Transformer-like (e.g. Llama) } & \makecell[l]{Perplexity, \\ Throughput} & \makecell[l]{QServe} &\makecell[l]{MIT EECS}  \\ \hline
        
        \makecell{LLMC} & \makecell[l]{AWQ, AdaDim, DGQ, GPTQ, \\FP8, HQQ, LLM.int8(), \\NormTweaking, OS+, OWQ, \\OmniQuant, QUIK, SmoothQuant, \\ SpQR, Quarot, Combinations} & \makecell[l]{Mixture-of-Expert, (e.g. Mixtral), \\ Transformer-like (e.g. Llama), \\ Multi-modal (e.g. LLaVA)} & \makecell[l]{Perplexity, \\ OpenCompass\footnotemark[36], \\ lm-evaluation-harness\footnotemark[37]} & \makecell[l]{TensorRT-LLM, \\MLC-TVM, vLLM, \\LightLLM,  Sglang, \\Lmdeploy, Transformers} &\makecell[l]{Beihang \& SenseTime}  \\ \hline
        
        \makecell{MI-optimize} & \makecell[l]{AWQ, FP8, GPTQ, QuIP, \\ RTN, SmoothQuant, SpQR, \\ ZeroQuant, Combinations} & \makecell[l]{Llama, Chatglm, Baichuan\\ } & \makecell[l]{BOSS (Robust), \\ Perplexity, \\lm-evaluation-harness\footnotemark[37]} & \makecell[l]{Transformers} &\makecell[l]{TsingmaoAI}  \\ \hline
        
        \makecell{QLLM-Eval} & \makecell[l]{AWQ, SmoothQuant} & \makecell[l]{Mixture-of-Expert, (e.g. Mixtral), \\ Transformer-like (e.g. Llama), \\ Multi-modal (e.g. LLaVA), \\Long-Context (e.g. Longchat), \\Others (e.g. Mamba)} & \makecell[l]{OpenCompass\footnotemark[36], \\ lm-evaluation-harness\footnotemark[37], \\ LongEval\footnotemark[38], Lost-in-the-middle\footnotemark[39],\\  MT-Bench\footnotemark[40]} & \makecell[l]{Transformers} &\makecell[l]{ Tsinghua University}  \\ \hline 

        LLM Compressor & \makecell[l]{GPTQ, SmoothQuant, FP8} & \makecell[l]{Mixture-of-Expert, (e.g. Mixtral) \\ Transformer-like (e.g. Llama) } & - & vLLM & Neural Magic \\ 
        
        
        \bottomrule
    \end{tabular}
    }
    \caption{Quantization toolkits and benchmarks for large language models. }
    \label{tab:sec2:quantization_toolkits}
\end{table*}
\footnotetext[36]{https://github.com/open-compass/opencompass}
\footnotetext[37]{https://github.com/EleutherAI/lm-evaluation-harness}
\footnotetext[38]{https://github.com/DachengLi1/LongChat}
\footnotetext[39]{https://github.com/nelson-liu/lost-in-the-middle}
\footnotetext[40]{https://github.com/lm-sys/FastChat/tree/main/fastchat/llm\_judge}



The quantization toolkits that are devoted to providing comprehensive comparisons have good support for the prevailing models and quantization algorithms in various aspects. Most toolkits include well-known models like the Llama series, Mixtral, Vicuna, and so on. Those who pay more attention to the model diversity, such as QLLM-Eval, have further support for various models. 
As well as the algorithms, LLMC, LMQuant, and MI-optimize focus on the performance of different quantization algorithms, and provide uniform, fair, comprehensive benchmarks for comparisons. 
All the benchmarks are based on one or several inference frameworks as backends, and leave interfaces for users to define and evaluate any custom models and algorithms easily. 


\subsubsection{Evaluation}
The evaluation in the benchmarks showcases the most interesting aspects of quantization LLMs, i.e., efficiency and generation quality. We list the detailed tracks in Table~\ref{tab:sec2:quantization_toolkits}. 
\textbf{For efficiency}, the inference efficiency is measured by deployability and throughput, which are the most crucial features in LLMs compression~\citep{lin2024qserve, gong2024llm}. Typically, reducing the storage of the parameters can speed up the inference theoretically, but it depends on the actual system implementation. The benchmark provides us with a fair and convenient probe to distinguish the algorithms and implementations that have practical acceleration and storage saving. 
The production efficiency is measured by calibration time, which indicates the time and computational resources cost of the PTQ algorithms~\citep{gong2024llm}. Methods that spare lots of resources usually have better generation quality, while those that require less time may have worse generation performance. It is a trade-off in producing quantized LLMs. 
\textbf{For generation quality}, it has many aspects, such as perplexity, accuracy, logic, completion, trustworthiness and so on~\citep{lin2024qserve, gong2024llm, li2024evaluating, liu2024evaluating}. Most benchmarks evaluate emergent capability, which is the key feature of LLMs. Specifically, models and algorithms are tested under diverse scenarios, like dialogue, long-context, or multi-task~\citep{li2024evaluating}. And some benchmarks are aware of the safety of generative contents, and estimate the trustworthiness and robustness of LLMs~\citep{li2024evaluating, liu2024evaluating}. 

\begin{tcolorbox}[
    colback=blue!5,  
    colframe=blue!50, 
    title={\textbf{Takeaways of \autoref{sec:toolkit}}}, 
    fonttitle=\bfseries,
    boxrule=1pt,
    arc=3pt,
]
\blue{
If your goal is to reproduce a variety of quantization algorithms, LLMC, MI-optimize, and LMQuant are recommended, as they provide a comprehensive suite of quantization methods. 
If your focus is to deploy across inference frameworks, LLMC stands out as an ideal choice, providing flexible quantization settings and seamless compatibility with multiple backends.
}
\end{tcolorbox}

\section{Future Trends and Directions}
\label{sec:future}

As the field of large language model quantization continues to evolve, several emerging trends and research directions are poised to shape its future. This section explores the anticipated advancements in quantization techniques, model architectures, and hardware design that will drive improvements in the efficiency, performance, and application of quantized models.

\textbf{Quantization Techniques.} Despite progress, several challenges remain in quantization techniques. Firstly, one major issue is the unclear  low-semantic-valueorigin of outliers in large language models (LLMs), which presents a significant barrier to further reducing quantization bit widths. Research aimed at uncovering the internal mechanisms behind these outliers is crucial and will provide valuable insights for the community, potentially advancing the state of quantization and enabling more efficient models. Secondly, pushing the boundaries of minimal bit representation with acceptable accuracy is highly valuable. Achieving the lowest possible bit width while maintaining performance can fully leverage hardware capabilities and maximize its potential. Thirdly, exploring unified strategies for mixed-bit quantization, including both bit selection and intra-layer/inter-layer bit allocation, is essential for optimizing model performance and efficiency. Current methods primarily emphasize intra-layer mixed precision, often overlooking the potential benefits of inter-layer mixed precision. Last but not least, developing semantic-guided strategies for achieving even lower-bit quantization and compression of key-value (KV) caches will be a major focus. During inference with long context lengths, the primary bottleneck often lies in the substantial memory usage of KV caches. Therefore, identifying effective methods for compressing KV caches is crucial for overcoming this limitation and enhancing model efficiency.

\textbf{Model Architecture.} Innovations in model architecture will also play a pivotal role. Firstly, quantizing models that handle multiple modalities will be explored to ensure efficiency across diverse data types and applications. Secondly, research will expand to include quantization strategies for new and emerging model structures such as the Mixture of Experts (MOE) and other large-scale architectures. Third, exploring the relationship between quantization and model size will provide insights into optimizing smaller models for performance while managing quantization trade-offs.

\textbf{Hardware Design.} Advancements in hardware and quantization co-design will be essential for unlocking new potential. The first area of focus is the development of systems for new types of extremely low-bit quantization. Innovative formats for low-bit representation and efficient system implementations may offer new solutions to the challenges posed by Moore's Law. The second area involves accelerating training with lower-bit precision, such as FP4. Research into hardware that supports training with such low-bit precision will be essential for speeding up model training while preserving performance.

\section{Conclusions}

In this survey, we have presented an in-depth exploration of low-bit quantization techniques for large language models (LLMs), highlighting their significance in addressing the computational and memory challenges associated with deploying these models in constrained environments. We began by elucidating the fundamentals of low-bit quantization, including the novel data formats and granularities that cater specifically to LLMs. Our review of systems and frameworks has illustrated the diverse approaches and tools available for supporting low-bit LLMs across different hardware platforms. We have also categorized and discussed various techniques for optimizing training and inference, providing a comprehensive understanding of current methodologies. Lastly, we have explored future directions and emerging trends in the field, emphasizing potential research areas and technological advancements that could further enhance the efficiency and effectiveness of LLM quantization. As the landscape of LLM research continues to evolve, this survey aims to serve as a valuable resource for advancing the development of low-bit quantization techniques.


\bibliographystyle{cas-model2-names}

\bibliography{cas-refs}


\end{document}